# Efficient Rotation-Scaling-Translation Parameters Estimation Based on Fractal Image Model

M. Uss[a], B. Vozel[b], V.Lukin[c], K. Chehdi[b]

[a] Department of Aircraft Radioelectronic Systems Design, National Aerospace University, 17 Chkalova St., Kharkov, 61070, Ukraine;

[b] IETR UMR CNRS 6164 - University of Rennes 1, CS 80518, 22305 Lannion cedex, France;

[c] Department of Transmitters, Receivers and Signal Processing, National Aerospace University, Kharkov, Ukraine;

M. L. Uss: phone: +38 050 3013333; e-mail: uss@xai.edu.ua.

B. Vozel: phone: +33 296469071; fax: +33 296469075; e-mail: benoit.vozel@univ-rennes1.fr.

V. V. Lukin: phone/fax: +38 057 3151186 ; email: lukin@ai.kharkov.com.

K. Chehdi: phone: +33 296469036; fax: +33 296469075; e-mail: kacem.chehdi@univ-rennes1.fr.

## ABSTRACT:

This paper deals with area-based subpixel image registration under rotation-isometric scaling-translation transformation hypothesis. Our approach is based on a parametrical modeling of geometrically transformed textural image fragments and maximum likelihood estimation of transformation vector between them. Due to the parametrical approach based on the fractional Brownian motion modeling of the local fragments texture, the proposed estimator $ML_{fBm}$ (ML stands for "Maximum Likelihood" and fBm for "Fractal Brownian motion") has the ability to better adapt to real image texture content compared to other methods relying on universal similarity measures like mutual information or normalized correlation. The main benefits are observed when assumptions underlying the fBm model are fully satisfied, e.g. for isotropic normally distributed textures with stationary increments. Experiments on both simulated and real images and for high and weak correlation between registered images show that the $ML_{fBm}$ estimator offers significant improvement compared to other state-of-the-art methods. It reduces translation vector, rotation angle and scaling factor estimation errors by a factor of about 1.75…2 and it decreases probability of false match by up to 5 times. Besides, an accurate confidence interval for $ML_{fBm}$ estimates can be obtained from the Cramér–Rao lower bound on rotation-scaling-translation parameters estimation error. This bound depends on texture roughness, noise level in reference and template images, correlation between these images and geometrical transformation parameters.

## 1. INTRODUCTION

Image registration is a fundamental image processing problem aiming at mapping two or more images (reference and template ones) to a common coordinate system [1]. Registration enables joint analysis of the information content of images acquired by different sensors at different time instances and/or under different modalities. Such practical and challenging use cases can be frequently met in remote sensing (registration of different spectral bands, images with large time-base gap between each other or different spatial/spectral resolutions, registration of optical and radar images) [2-4] or in medical imaging (registration of computed tomography, magnetic resonance, and photon emission tomography images) [5].

A large number of image registration methods often determine parameters of a global geometrical transformation between reference and template images using a set of linked control fragments (CF). By CF, let us mean here a small image fragment with a practically similar content recognizable in both images (for feature-based methods, Control Points or Feature Points terms are in use). In practice, these CFs have to be selected first in both images and they can be subsequently registered either by feature-based or by area-based methods [3, 4]. In the former case, a rather large initial image registration error is tolerated provided the time required for finding and linking the CFs is limited and reasonable. On the contrary, area-based algorithms put more emphasis on the achievable CFs registration accuracy accepting thus higher computational complexity [6]. As a result, feature-based methods have found a wide use at the coarse registration stage whilst area-based methods are often preferred at the fine registration stage, especially when subpixel registration accuracy is required [7, 8].

Without loss of generality, the area-based registration problem aims at obtaining an accurate estimation of geometrical transformation parameters between two CFs (or a couple of small reference and template image fragments) relying directly on pixel intensities in these fragments. Due to the rigorous positioning of modern satellite sensors on one hand and the local nature of the problem at a CF level on the other hand, linear geometrical transform models between the two CFs can be reasonably considered [2], such as pure translation, rotation-scaling-translation (RST) or affine transformation [9] to name the most commonly used. In some cases, a correction for the relief influence might be required in addition to the previous assumption on the geometrical transform model. In this paper, we essentially concentrate on RST transformation model with isometric scaling between the two CFs.

Area-based registration can be viewed as an optimization problem of a suitable similarity measure between reference and template CFs. There are few widespread similarity measures. The simplest one is sum of squared differences (SSD) [3]. This distance measure implicitly assumes that the intensity values of the corresponding fragments in two registered images are more or less within the same magnitude order. The use of this distance measure can certainly provide correct results when the aforementioned hypothesis is strictly satisfied. Otherwise, the results may degrade, in particular for multimodal images. The cross-correlation or least squares similarity measure can be viewed as an extension for handling linear dependence between the reference and template images intensities [8]. In multimodal settings, a standard solution is to consider a normalized version of the cross-correlation (Normalized Cross-Correlation, NCC) [10]. NCC is, arguably, the most frequently used similarity measure in image registration [11]. It is the basis for the Correlation- and Hough transform-based method of Automatic Image Registration (CHAIR) approach recently proved to cope with complex registration cases including synthetic aperture radar (SAR) with optical images registration [2].

The mutual information (MI) distance measure, such as the one introduced in [12, 13] for registration, allows tackling with even more complex dependence between the reference and template images. The underlying idea is to measure the normalized entropy of joint density of the reference and template images. A Parzen-window estimator [14] with a smooth compactly supported kernel function can be used for estimating the unknown joint density.

The normalized image intensity gradients (Normalized Gradient Fields, NGF) method [15] achieves a compromise between the more restricted SSD and the very general (and highly nonconvex) MI. This measure assumes that intensity changes in images of different modalities appear at corresponding positions. It is basically an L2-norm of a residual, measuring the alignment of the normalized gradients of reference and template images at a given position. Normalization of the gradient allows focusing on locations of changes rather than on the strength of the changes.

Within this framework, subpixel registration accuracy can be usually achieved using interpolation of reference or/and template images [11]. This additional stage might have negative effect on geometrical transformation parameters estimation accuracy (for example, introducing bias) as it is discussed in [16, 17]. All the abovementioned similarity measures were adopted in the past quite successfully to measure either pure translation [11], or RST parameters [18], or more complex geometrical transformations model parameters [8, 19] with subpixel accuracy.

In multitemporal and /or multimodal case, it happens that correlation between reference and template CFs may tend to be moderate or even weak; strongly correlated CFs could be rather rare in a pair of images to register. In such specific conditions, a registration method should be able to use available data as effectively as possible. More strictly, it should be characterized by a high probability of positive match and high registration accuracy in a wide range of correlation between reference and template images – from strong to weak. However, despite the research efforts devoted towards achieving this goal, design of registration methods with such wide application spectrum is still an open problem.

In particular, the methods based on universal measures such as SSD, NCC, MI or NGF cannot meet the abovementioned requirement easily. They impose only general requirements on registered images like smoothness or statistical dependence. They do not implicitly take into account image content and/or noise statistics. Such drawback inevitably reduces registration efficiency.

Additionally, in multitemporal and/or multimodal registration cases, it is a difficult problem to precisely quantify the final accuracy of estimated parameters for a given geometrical transformation. The two main reasons for this lie in a rather complex structure of similarity measures in general and the often negative influence of interpolation stage. A Cramér–Rao lower bound (CRLB) on translation estimation error based on SSD measure was obtained by D. Robinson and P. Milanfar in [20]. This work was further extended for 2D rotation, RST transformation, 2D and 3D affine transformations [21] and 2D projective transformation [22]. As it has been shown in [23], this bound can be rather inaccurate in describing real estimators' performance. Besides, it cannot be directly applied to multitemporal and/or multimodal cases.

In a recent paper [23], we proposed and studied an original CRLB on pure translation estimation error STD. This bound was experimentally compared to other similar bounds of the literature. The performance of standard translation estimators was compared against these set of bounds based on simulated and real data. The obtained results showed good accuracy and adequateness of the newly proposed bound in a variety of settings including multitemporal and/or multimodal cases. A significant gap between theoretically predicted accuracy and performance of real registration methods has thus been filled to a certain extent in the case of simple translation transformation.

In this paper, we move forward and prove it is possible to reduce further this gap, that is, to derive a new registration method that performs closer to the theoretically predicted accuracy for both a restricted but wide enough class of images and more complex geometrical transformations. A new and very efficient area-based registration method is thus proposed and its accuracy is precisely quantified.

First of all, we upgrade the geometrical transformation model considered in our approach, from a simple translation to a more complex and more realistic RST transformation model, better suited for many remote sensing applications.

Second, in contrast to our previous work, we focus hereafter more properly on the efficient estimation of RST parameters in multitemporal and/or "mild" multimodal settings (registering, in particular, real images acquired by VNIR and SWIR optical sensors in the experimental part of this paper). We consequently propose a new estimator within the same approach as in [23] but extended to RST transformation hypothesis between the two CFs to register. It is called $ML_{fBm}$ due to its two distinctive features: it is derived within the Maximum Likelihood framework and the local texture in both CFs is still assumed to be well modeled by the fractal Brownian motion model (in $ML_{fBm}$ "ML" stands for "Maximum Likelihood" and "fBm" for "fractal Brownian motion"). The $ML_{fBm}$ estimator is proposed along with a refined optimization scheme assuring its global convergence.

More, the observation model considered in this new $ML_{fBm}$ estimator relies on a signal-dependent noise model for both RI and TI. This model proved to be more adequate for new generation of multispectral and hyperspectral sensors [24, 25]. At the CF level, we approximate the assumed signal-dependent noise by an additive noise with signal-dependent variance. This distinctive feature of the $ML_{fBm}$ method is worth noticing as other area-based registration methods cannot implicitly take such noise properties into account. We show a quite large tolerance of the $ML_{fBm}$ estimator to errors in the noise variance. This property allows using a possibly inaccurate noise variance directly estimated from noisy images without facing a decrease of the proposed method efficiency (meaning that no *a priori* information on noise variance is required to operate safely the method).

Besides, we show that the $ML_{fBm}$ estimator is able to reduce RST parameters estimation error by a factor of 1.75…2 as compared to state-of-the-art methods. The $ML_{fBm}$ method is also characterized by a significantly lower probability of false CFs registration (outlier occurrence among CFs). It can deal in practice with large temporal and spectral differences, different spatial resolutions of reference and template images, weak correlation between registered CFs (normalized correlation coefficient down to 0.4 is acceptable). Effects of relief influence can also be taken into account with the $ML_{fBm}$ using digital elevation model (DEM).

An extra outcome of the work performed in this paper is a CRLB describing the potential accuracy of parameters estimation for RST transformation hypothesis between couples of CPs. This CRLB is used here to assign a confidence interval for the obtained RST parameters estimates (confidence ellipsoid based on a CRLB estimate). To the best of our knowledge, this is the only bound of this kind suitable for multitemporal and/or multimodal registration cases. This bound can be especially useful to assign weights to each RST parameter estimate depending on its actual accuracy. Such weights can be later used either for outlier detection or for obtaining an adequate weighted estimate of the global geometrical transformation parameters.

Finally, we investigate experimentally the range of applicability of the proposed parametrical approach to real data and demonstrate that for rather wide image class including isotropic textures with normal increments, the proposed method is more efficient than state-of-the-art methods and perform very close to the corresponding CRLB.

The paper is organized as follows. Section 2 introduces the parametric statistical model chosen for describing translated, mutually rotated and scaled image textures and it details the $ML_{fBm}$ estimator developed accordingly. In Section 3, the performance of the newly proposed $ML_{fBm}$ estimator is comparatively assessed against that of four other alternative estimators based on experiments on simulated pure fBm data. The performance of this $ML_{fBm}$ estimator is analyzed in Section 4 for real-life Hyperion and Landsat 8 data. Finally, discussion and conclusions are given in Sections 5 and 6.

## 2. JOINT MAXIMUM LIKELIHOOD ESTIMATION OF RST TRANSFORMATION AND IMAGE TEXTURE PARAMETERS

This Section formally defines the newly derived $ML_{fBm}$ estimator of RST transformation parameter vector between reference and template images control fragments. Its potential performance characteristics are analyzed and convergence issues are discussed.

### 2.1. Problem statement

By reference/template CF we mean image fragments of small size (from 7 by 7 to about 25 by 25 pixels) cut out from the full size reference/template images. They are defined at two local reference/template coordinate systems with axes $tO_{RI}s/uO_{TI}v$, where $(t,s)$ and $(u,v)$ denote respective pixel coordinates, and origins $O_{RI}$ and $O_{TI}$ are placed in the center of the corresponding CFs. In what follows, we use subscripts “RI” and “TI” for reference and template CFs, respectively. “XX” stands for either “RI” or “TI” according to the context.

We assume the RST transformation model between $tO_{RI}s$ and $uO_{TI}v$ coordinate systems that includes rotation by an angle $\alpha$, isometric scaling with a factor $\Delta r$ and translation by a vector $(\Delta t, \Delta s)$, where $\Delta t$ and $\Delta s$ are vertical and horizontal translation components:

$$\begin{pmatrix} u \\ v \end{pmatrix} = \Delta r \begin{pmatrix} \cos\alpha & \sin\alpha \\ -\sin\alpha & \cos\alpha \end{pmatrix} \begin{pmatrix} t \\ s \end{pmatrix} + \begin{pmatrix} \Delta t \\ \Delta s \end{pmatrix}. \quad (1)$$

The RST model parameter vector $\boldsymbol{\theta}_{\mathrm{RST}} = (\Delta t, \Delta s, \alpha, \Delta r)$ is to be estimated with accuracy allowing subpixel alignment of reference and template CFs.

The reference, $y_{\mathrm{RI}}(t,s)$, and template, $y_{\mathrm{TI}}(u,v)$, CFs are of size $N_{\mathrm{RI}} \times N_{\mathrm{RI}}$ and $N_{\mathrm{TI}} \times N_{\mathrm{TI}}$ pixels respectively. They are defined according to the following additive observation model:

$$y_{\mathrm{RI}}(t,s) = x_{\mathrm{RI}}(t,s) + \eta_{\mathrm{RI}}(t,s),\ t = -N_{h.\mathrm{RI}},\dots,N_{h.\mathrm{RI}}, s = -N_{h.\mathrm{RI}},\dots,N_{h.\mathrm{RI}},$$

$$y_{\mathrm{TI}}(u,v) = x_{\mathrm{TI}}(u,v) + \eta_{\mathrm{TI}}(u,v),\ u = -N_{h.\mathrm{TI}},\dots,N_{h.\mathrm{TI}}, \mathrm{v} = -N_{h.\mathrm{TI}},\dots,N_{h.\mathrm{TI}},$$

where $N_{h.XX} = (N_{XX} - 1)/2$, $x_{\mathrm{RI}}(t,s)$ and $x_{\mathrm{TI}}(u,v)$ are pixel samples of the RI and TI noise-free CFs, respectively; $\eta_{\mathrm{RI}}(t,s)$ and $\eta_{\mathrm{TI}}(u,v)$ are the corresponding noise processes viewed as stationary, spatially uncorrelated, zero-mean, Gaussian distributed fields with variances $\sigma^2_{n.\mathrm{RI}}$ and $\sigma^2_{n.\mathrm{TI}}$, respectively, and independent of each other. With these definitions, $N_{\mathrm{RI}}$ and $N_{\mathrm{TI}}$ are considered as odd values in our work. To deal with a signal-dependent noise hypothesis, $\sigma^2_{n.\mathrm{RI}}$ and $\sigma^2_{n.\mathrm{TI}}$ are allowed to vary from CF to CF as this situation will be described in subsection 4.1. Other choices of CF shape (non-symmetrical arbitrary shape) are possible without modifying the proposed method.

Both $y_{\text{RI}}(t,s)$ and $y_{\text{TI}}(u,v)$ are transformed into $N_{XX}^2 \times 1$ column vectors $\mathbf{Y}_{\text{RI}}$ and $\mathbf{Y}_{\text{TI}}$ in column-major order and they compose sample $\mathbf{Y}_{\Sigma} = \begin{pmatrix} \mathbf{Y}_{\text{RI}} \\ \mathbf{Y}_{\text{TI}} \end{pmatrix}$ of size $(N_{\text{RI}}^2 + N_{\text{TI}}^2) \times 1$. Let us define coordinates of a *k*-th element of $\mathbf{Y}_{\text{RI}}$ vector as $(t_k, s_k)$, and an *l*-th element of $\mathbf{Y}_{\text{TI}}$ vector as $(u_l, v_l)$.

We adopt fBm model [26] to locally describe image texture or, more precisely, obtain correlation matrix of the sample $\mathbf{Y}_{\Sigma}$. A great advantage of considering the fBm model for characterizing local texture is that it allows describing complex shapes with only two parameters [27]: texture roughness parameterized by Hurst exponent $H$ and texture amplitude parameterized by $\sigma_x$. Here $H \in [0,1]$ (values less than 0.5 corresponds to rough and greater than 0.5 - to smooth textures) and $\sigma_x$ is standard deviation (STD) of texture increments on unit distance. We additionally assume the same value of the parameter $H$ for both reference and template images (later in subsection 4.5, we have checked that this assumption is justified for real data). Thus, the registration problem is parameterized with only eight parameters (low order models are known to be preferable when estimating parameters from small samples [28, 29]) forming the full parameter vector $\boldsymbol{\theta} = (\sigma_{x.\text{RI}}, \sigma_{x.\text{TI}}, H, k_{\text{RT}}, \Delta t, \Delta s, \alpha, \Delta r)$, where $\sigma_{x.\text{RI}}, \sigma_{x.\text{TI}}$ are $\sigma_x$ values for reference and template CFs, respectively, $k_{\text{RT}}$ is the correlation coefficient between these pair of CFs. Noise variances $\sigma_{n.XX}^2$ are supposed to be known and, accordingly, they are not included in $\boldsymbol{\theta}$ (in practice, $\sigma_{n.XX}^2$ can be found either using a sensor calibration dataset or estimated directly based on the image data, as suggested in our recent works [30, 31]). The full parameter vector can be represented as $\boldsymbol{\theta} = \left(\boldsymbol{\theta}_{\text{texture}}, \boldsymbol{\theta}_{\text{RST}}\right)$, where $\boldsymbol{\theta}_{\text{texture}} = (\sigma_{x.\text{RI}}, \sigma_{x.\text{TI}}, H, k_{\text{RT}})$ is the texture parameter vector and $\boldsymbol{\theta}_{\text{RST}}$ is the RST parameter vector defined above. Within the framework introduced above, image registration problem amounts to estimating the vector $\boldsymbol{\theta}$.

By using a local parametric approach for solving the registration problem, we seek to increase the final registration efficiency by a better adaptation of the whole approach to image texture. However, the fBm model is suitable for describing isotropic normally distributed textures with stationary increments. Thus, the proposed method needs to be used cautiously when Gaussian hypothesis (single-look SAR images with fully developed speckle is such an example) and/or isotropy hypothesis on local texture or noise model violates. We consider this issue more in detail later in Section 4.

### 2.2. The $\mathbf{ML_{fBm}}$ estimator

Let us now introduce the maximum likelihood estimator (MLE) of the vector $\boldsymbol{\theta} = (\boldsymbol{\theta}_{\text{texture}}, \boldsymbol{\theta}_{\text{RST}})$. According to the definition of fBm process, it takes zero value at the origin. To assure this, consider a new sample $\Delta\mathbf{Y}_\Sigma = \begin{pmatrix} \Delta\mathbf{Y}_{\text{RI}} \\ \Delta\mathbf{Y}_{\text{TI}} \end{pmatrix} = \begin{pmatrix} \mathbf{Y}_{\text{RI}} - x_{\text{RI.0}}\mathbf{1}_{\text{RI}} \\ \mathbf{Y}_{\text{TI}} - x_{\text{TI.0}}\mathbf{1}_{\text{TI}} \end{pmatrix}$, where $x_{\text{XX.0}} = x_{XX}(0,0)$ denotes true values of reference/template CFs central pixel, $\mathbf{1}_{\text{XX}}$ are unit vectors of size $N_{XX}^2 \times 1$, respectively. The correlation matrix $\mathbf{R}_\Sigma$ of the sample $\Delta\mathbf{Y}_\Sigma$ is given by [23]:

$$\mathbf{R}_\Sigma = \begin{pmatrix} \mathbf{R}_{\text{RI}} + \mathbf{R}_{n.\text{RI}} & k_{\text{RT}} \cdot \mathbf{R}_{\text{RT}}(\boldsymbol{\theta}_{\text{RST}}) \\ k_{\text{RT}} \cdot \mathbf{R}_{\text{RT}}^T(\boldsymbol{\theta}_{\text{RST}}) & \mathbf{R}_{\text{TI}} + \mathbf{R}_{\text{n.TI}} \end{pmatrix}, \tag{2}$$

where $\mathbf{R}_{XX}$ are correlation matrices of noise-free $\Delta\mathbf{Y}_{XX}$ sample, $k_{\text{RT}}\mathbf{R}_{\text{RT}}$ is the cross-correlation matrix between $\Delta\mathbf{Y}_{\text{RI}}$ and $\Delta\mathbf{Y}_{\text{TI}}$, $\mathbf{R}_{n.\text{XX}} = \sigma_{n.XX}^2 \mathbf{I}_{XX}$ are correlation matrices of noise for reference/template CFs, $\mathbf{I}_{XX}$ are $N_{XX} \times N_{XX}$ identity matrices, respectively.

Let $\mathbf{R}_{\text{RT}}$ be expanded as $\mathbf{R}_{\text{RT}} = \sigma_{x.\text{RI}}\sigma_{x.\text{TI}}\mathbf{R}_{\text{HRT}}$. Here elements of the matrix $\mathbf{R}_{\text{HRT}}$ describe covariance between elements of $\Delta\mathbf{Y}_{\text{RI}}$ and $\Delta\mathbf{Y}_{\text{TI}}$ when $\sigma_{x.\text{RI}} = \sigma_{x.\text{TI}} = 1$ and $k_{\text{RT}} = 1$. For the fBm model, elements $R_{\text{RI}}(k_1,k_2)$, $R_{\text{TI}}(l_1,l_2)$, and $R_{\text{HRT}}(k,l)$ take the following form (see Appendix A for details):

$$R_{\text{RI}}(k_1,k_2) = 0.5\sigma_{x.\text{RI}}^2 \left[ \left(t_{k_1}^2 + s_{k_1}^2\right)^H + \left(t_{k_2}^2 + s_{k_2}^2\right)^H - \left( \left(t_{k_1} - t_{k_2}\right)^2 + \left(s_{k_1} - s_{k_2}\right)^2 \right)^H \right],$$

$$R_{\text{TI}}(l_1,l_2) = 0.5\sigma_{x.\text{TI}}^2 \left[ \left(u_{l_1}^2 + v_{l_1}^2\right)^H + \left(u_{l_2}^2 + v_{l_2}^2\right)^H - \left( \left(u_{l_1} - u_{l_2}\right)^2 + \left(v_{l_1} - v_{l_2}\right)^2 \right)^H \right],$$

$$R_{\text{HRT}}(k,l) = \frac{\Delta r^H}{2} \left[ \left( \left(t_k - t_0'\right)^2 + \left(s_k - s_0'\right)^2 \right)^H + \left(t_l'^2 + s_l'^2\right)^H - \left(t_0'^2 + s_0'^2\right)^H - \left( \left(t_k - t_l'\right)^2 + \left(s_k - s_l'\right)^2 \right)^H \right],$$

where $t' = \Delta r^{-1}\left(\cos\alpha(u - \Delta t) - \sin\alpha(v - \Delta s)\right)$, $s' = \Delta r^{-1}\left(\sin\alpha(u - \Delta t) + \cos\alpha(v - \Delta s)\right)$,

$t_0^{'} = -\Delta r^{-1}\left(\Delta t\cos\alpha - \Delta s\sin\alpha\right)$, $s_0^{'} = -\Delta r^{-1}\left(\Delta t\sin\alpha + \Delta s\cos\alpha\right)$.

Omitting a constant that does not depend on $\boldsymbol{\theta}$, the logarithmic likelihood function (log-LF) of the sample $\Delta\mathbf{Y}_\Sigma$ can be written as:

$$\log L\left(\Delta\mathbf{Y}_\Sigma,\boldsymbol{\theta}\right) = -\frac{1}{2}\left(\Delta\mathbf{Y}_\Sigma{}^T\mathbf{R}_\Sigma^{-1}\Delta\mathbf{Y}_\Sigma + \log\left|\mathbf{R}_\Sigma\right|\right). \quad (3)$$

With these notations, the MLE of the parameter vector $\boldsymbol{\theta}$ is obtained as:

$$\hat{\boldsymbol{\theta}} = \arg\max_{\boldsymbol{\theta}}\left[\log L\left(\Delta\mathbf{Y}_\Sigma;\boldsymbol{\theta}\right)\right], \quad (4)$$

$$\begin{pmatrix}\hat{x}_{\mathrm{RI.0}}\\ \hat{x}_{\mathrm{TI.0}}\end{pmatrix} = \begin{pmatrix}\mathbf{e}_{\mathrm{RI}}^T\mathbf{R}_\Sigma^{-1}\mathbf{e}_{\mathrm{RI}} & \mathbf{e}_{\mathrm{RI}}^T\mathbf{R}_\Sigma^{-1}\mathbf{e}_{\mathrm{TI}}\\ \mathbf{e}_{\mathrm{TI}}^T\mathbf{R}_\Sigma^{-1}\mathbf{e}_{\mathrm{RI}} & \mathbf{e}_{\mathrm{TI}}^T\mathbf{R}_\Sigma^{-1}\mathbf{e}_{\mathrm{TI}}\end{pmatrix}^{-1}\begin{pmatrix}\mathbf{e}_{\mathrm{RI}}^T\mathbf{R}_\Sigma^{-1}\Delta\mathbf{Y}_\Sigma\\ \mathbf{e}_{\mathrm{TI}}^T\mathbf{R}_\Sigma^{-1}\Delta\mathbf{Y}_\Sigma\end{pmatrix}, \quad (5)$$

subject to constraints $\sigma_{x.\mathrm{RI}} \geq 0$; $\sigma_{x.\mathrm{TI}} \geq 0$; $0 \leq H \leq 1$; $\left|k_{\mathrm{RT}}\right| \leq 1$. Here $\mathbf{e}_{\mathrm{RI}} = \left(\mathbf{1}_{\mathrm{RI}},\mathbf{0}_{\mathrm{TI}}\right)$ and $\mathbf{e}_{\mathrm{TI}} = \left(\mathbf{0}_{\mathrm{RI}},\mathbf{1}_{\mathrm{TI}}\right)$, $\mathbf{0}_{XX}$ are $N_{XX}^2 \times 1$ zero vectors. MLE of unknown values $x_{\mathrm{XX.0}}$ in (5) are obtained by equating to zero the first derivatives of $\log L\left(\Delta\mathbf{Y}_\Sigma,\boldsymbol{\theta}\right)$ w.r.t. $x_{\mathrm{XX.0}}$.We will later refer to the estimator in (4) as the newly proposed $\mathrm{ML_{fBm}}$ estimator for RST geometrical transformation hypothesis.

The $\mathrm{ML_{fBm}}$ estimator in (4) optimizes the similarity measure (3). It is important to note that the log-LF in (3) is a continuous function w.r.t. the parameters vector $\boldsymbol{\theta}$ and does not involve any transformation of the input data $\Delta\mathbf{Y}_\Sigma$. Therefore, subpixel registration accuracy can be reached with $\mathrm{ML_{fBm}}$ estimator without interpolating either image data or similarity measure. This is a positive feature of the $\mathrm{ML_{fBm}}$ worth noticing as it has been repeatedly emphasized in the literature that such interpolation stage might alter accuracy of subpixel registration algorithms [16, 17, 20].

By using the $\mathrm{ML_{fBm}}$ estimator, the lower bound on estimation error STD of parameter vector $\boldsymbol{\theta}$ can be calculated as:

$$\boldsymbol{\sigma}_{\boldsymbol{\theta}} = \sqrt{diag(\mathbf{C}_{\boldsymbol{\theta}})}, \quad (6)$$

where $diag(\cdot)$ returns the diagonal elements of a matrix, $\mathbf{C}_{\boldsymbol{\theta}} = \mathbf{I}_{\boldsymbol{\theta}}^{-1}$ is the CRLB on estimation errors covariance, $\mathbf{I}_{\boldsymbol{\theta}}$ is the Fisher Information Matrix (FIM) of the parameter vector $\boldsymbol{\theta}$ with ij-th entry [28]:

$$\mathbf{I}_{\boldsymbol{\theta}}(i,j) = I_{\boldsymbol{\theta}(i)\boldsymbol{\theta}(j)} = \frac{1}{2}\mathrm{tr}\left(\frac{\partial \mathbf{R}_\Sigma}{\partial \boldsymbol{\theta}(i)}\mathbf{R}_\Sigma^{-1}\frac{\partial \mathbf{R}_\Sigma}{\partial \boldsymbol{\theta}(j)}\mathbf{R}_\Sigma^{-1}\right),\ i,j = 1...8. \quad (7)$$

Derivatives of $\mathbf{R}_\Sigma$ w.r.t. $\boldsymbol{\theta}(i)$ are given in Appendix A. We denote by $\boldsymbol{\sigma}_{\mathrm{RST}}$ ($\mathbf{C}_{\boldsymbol{\theta}_{\mathrm{RST}}}$) the part of $\boldsymbol{\sigma}_{\boldsymbol{\theta}}$ ($\mathbf{C}_{\boldsymbol{\theta}}$) related to RST parameters $\Delta t$, $\Delta s$, $\alpha$ and $\Delta r$ defined above.

Given matrix $\mathbf{C}_{\boldsymbol{\theta}}$, a confidence interval on the MLE $\hat{\boldsymbol{\theta}}$ can be represented by the scattering ellipse in the parameters space. Therefore, the $\mathrm{ML_{fBm}}$ estimator can be also viewed as an interval estimator of the RST parameters. Accuracy of the interval estimates provided by the $\mathrm{ML_{fBm}}$ estimator depends on the actual adequacy of $\mathbf{C}_{\boldsymbol{\theta}}$ bound. A detailed analysis of $\mathbf{C}_{\boldsymbol{\theta}}$ for pure translation model [23] proved it to be a very tight bound even when dealing with real data. We will show in the next two Sections that this statement can be also extended to the RST model. To the best of our knowledge, our bound is the only one that can be applied at the moment to multitemporal and/or multimodal registration problem.

### 2.3. $\mathrm{ML_{fBm}}$ estimator initialization and implementation

The problem defined in (4) is a nonlinear constrained optimization problem and it is solved here using Han-Powell optimization method [32]. Advantages of this quasi-Newton method are superlinear convergence speed and availability of efficient implementations.

However, the log-LF given in (3) exhibits multiple extrema (see subsection 4.4). Therefore, a proper selection of an initial guess for $\hat{\boldsymbol{\theta}}$ is needed to prevent numerical optimization process from possible convergence to a local extremum. By definition, $\sigma_x^2$ is the variance of fBm-field increments on unit distance. Then, reasonable initial guesses for $\sigma_{x.\mathrm{RI}}$ and $\sigma_{x.\mathrm{TI}}$ can be obtained as standard deviation (STD) of $\mathbf{Y}_{\mathrm{XX}}$ first-order increments:

$$\hat{\sigma}_{x.RI}^2 = \left[D\left(y_{RI}(t,s) - y_{RI}(t+1,s)\right) + D\left(y_{RI}(t,s) - y_{RI}(t,s+1)\right)\right]/2, \quad (8)$$

$$\hat{\sigma}_{x.TI}^2 = \left[D\left(y_{TI}(u,v) - y_{TI}(u+1,v)\right) + D\left(y_{TI}(u,v) - y_{TI}(u,v+1)\right)\right]/2, \quad (9)$$

where the operator $D(\cdot)$ returns argument variance. We fix the initial guess for the Hurst parameter to $H = 0.5$, i.e. in the middle of the Hurst exponent range of possible values. The sample correlation coefficient between reference and template images is used as initial guess for $k_{\mathrm{RT}}$.

Setting initial guess for $\boldsymbol{\theta}_{\mathrm{RST}}$ vector depends on a particular application. Our recommendation for satisfying global convergence will be discussed in Section 4.

## 3. Comparative analysis of the $\mathrm{ML}_{\mathrm{fBM}}$ estimator against state-of-the-art alternatives on simulated fBm data

To better analyze the $\mathrm{ML}_{\mathrm{fBm}}$ ability to improve RST parameters estimation accuracy, let us first compare it against the most commonly used area-based similarity measures introduced above, such as the SSD [3], NCC [33], MI [12, 13], and NGF [15]. In this Section, comparison is carried out in controlled conditions based on simulated noisy fBm texture. All estimators are compared in terms of bias, efficiency (closeness to the $\mathbf{C}_{\boldsymbol{\theta}}$ bound), and distribution of RST parameters estimates.

Experimental results presented in this Section have been obtained based on the Flexible Algorithms for Image Registration (FAIR) software [34], a package written in MATLAB.

### 3.1. Test points

The following analysis is based on ten different test points (TP) numbered from 1 to 10 in Table 1. Among these test points (sets of parameters), TP #1 is treated as a basic parameter vector. The nine other TPs are obtained by changing one or several parameter value(s) of TP #1 components (those marked by bold in Table 1). TPs ##1…10 cover situations with rough and smooth texture, low and high noise level, weak and strong correlation between reference and template CFs (see the column Description).

We would like to stress that values of fBm model and RST parameters for the selected set of TPs in Table 1 are typical ones estimated for Landsat8 to Hyperion images registration problem discussed later in the experimental Section of this paper. The most frequently met value of the ratio $\sigma_x / \sigma_n$ for both Hyperion and Landsat8 bands is about 5 and it can drop down to 1 for noisy areas. The average of the Hurst exponent is about 0.65 but can be as low as 0.3 for some CFs; $|k_{\mathrm{RT}}|$ varies from 0 to 0.95 and we set $k_{\mathrm{RT}} = 0.95$ and 0.5 as the strong and weak correlation cases, respectively;

$(\Delta t, \Delta s)$ pairs cover subpixel shifts from no translation to half-pixel translation cases (integer shifts were removed from consideration here as they do not affect estimators performance). Rotation angle between Hyperion and Landsat8 images was about 17° and the scaling factor was about 1.025.

Table 1. Test points parameter values ($\sigma_{n.\mathrm{TI}} = \sigma_{n.\mathrm{RI}} = 1$, $\sigma_{x.\mathrm{RI}} = 5$, $N_{\mathrm{RI}} = N_{\mathrm{TI}} + 8$)

| Test point | Description | $\sigma_{x.\mathrm{TI}}$ | $H$ | $k_{\mathrm{RT}}$ | $N_{\mathrm{TI}}$ | $\Delta t$, pixels | $\Delta s$, pixels | $\alpha$, degrees | $\Delta r$ |
|---|---|---|---|---|---|---|---|---|---|
| 1 | Basic | 5 | 0.65 | 0.95 | 15 | 0.25 | 0.25 | 17 | 1.025 |
| 2 | Weak correlation | 5 | 0.65 | **0.5** | 15 | 0.25 | 0.25 | 17 | 1.025 |
| 3 | Small template CF size | 5 | 0.65 | 0.95 | **9** | 0.25 | 0.25 | 17 | 1.025 |
| 4 | High noise level | **1** | 0.65 | 0.95 | 15 | 0.25 | 0.25 | 17 | 1.025 |
| 5 | Rough texture | 5 | **0.35** | 0.95 | 15 | 0.25 | 0.25 | 17 | 1.025 |
| 6 | Pure translation | 5 | 0.65 | 0.95 | 15 | **0.5** | **0.5** | **0** | **1** |
| 7 | Pure translation | 5 | 0.65 | 0.95 | 15 | **0.5** | **0** | **0** | **1** |
| 8 | Pure rotation | 5 | 0.65 | 0.95 | 15 | **0** | **0** | **5** | **1** |
| 9 | Pure scaling | 5 | 0.65 | 0.95 | 15 | **0** | **0** | **0** | **0.8** |
| 10 | Zero geometrical transformation | 5 | 0.65 | 0.95 | 15 | **0** | **0** | **0** | **1** |

CRLBs on RST parameters estimation error STD for TP ##1…10 are given in Table 2. These values are calculated by substituting the corresponding parameters into Eq. (6) and (7). The lowest estimation accuracy is observed for TP #2 due to weak correlation between reference and template CFs, the highest – for TP #9. Mean theoretical estimation error STD is about 0.067 pixels for translation, 0.67° for rotation angle and about 0.012 for scaling factor.

Table 2. CRLB on RST parameters estimation error for TP ##1…10 (STD values)

| TP | $\boldsymbol{\sigma}_{\mathrm{RST}}(1)$, pixels ($\Delta t$) | $\boldsymbol{\sigma}_{\mathrm{RST}}(2)$, pixels ($\Delta s$) | $\boldsymbol{\sigma}_{\mathrm{RST}}(3)$, degrees ($\alpha$) | $\boldsymbol{\sigma}_{\mathrm{RST}}(4)$ ($\Delta r$) |
|---|---|---|---|---|
| 1 | 0.048 | 0.049 | 0.447 | 0.008 |
| 2 | 0.130 | 0.133 | 1.208 | 0.023 |
| 3 | 0.082 | 0.083 | 1.236 | 0.024 |
| 4 | 0.107 | 0.109 | 0.990 | 0.019 |
| 5 | 0.058 | 0.062 | 0.569 | 0.010 |
| 6 | 0.056 | 0.056 | 0.509 | 0.009 |
| 7 | 0.043 | 0.068 | 0.476 | 0.009 |
| 8 | 0.049 | 0.049 | 0.45 | 0.010 |
| 9 | 0.039 | 0.034 | 0.373 | 0.003 |
| 10 | 0.049 | 0.049 | 0.454 | 0.008 |

### 3.2. Numerical results analysis

For each test point, the reference and template CFs are obtained via Cholesky decomposition of the correlation matrix $\mathbf{R}_{\Sigma}$ [35]. A total number of 1000 samples is used to collect statistics for each estimator compared. Note that the reference CF size is set larger than the template CF size to ensure full overlapping of control fragments.

Several quantitative criteria are considered for assessing the estimation accuracy. We have decided to use median and median of absolute deviations (MAD) measures to account for possible outliers among estimates. For each $i$th component of the RST parameter vector, the quantitative criteria are defined by the following expressions: robust analogs of bias $b_i = \boldsymbol{\theta}_{\mathrm{RST}}(i) - med(\hat{\boldsymbol{\theta}}_{\mathrm{RST}}(i))$ and standard deviation $s_i = 1.48 \cdot MAD_i$, the statistical efficiency measure $e_i = 100\% \cdot \boldsymbol{\sigma}_{\mathrm{RST}}(i)^2 / MSE(i)$, $i = 1...4$. Here $med(\cdot)$ denotes median operator, $MAD_i = med(|\hat{\boldsymbol{\theta}}_{\mathrm{RST}}(i) - med(\hat{\boldsymbol{\theta}}_{\mathrm{RST}}(i))|)$ is median absolute deviation, $MSE(i) = s_i^2 + b_i^2$ is mean square error (for biased estimates), $e_i$ reflects efficiency of each estimator w.r.t. the $\boldsymbol{\sigma}_{\mathrm{RST}}(i)$ bound. For an efficient estimator, $e_i \approx 100\%$. A value $e_i \ll 100\%$ relates to a non efficient estimator.

Comparative results are presented in Fig. 1, 2 and Table 3. Recall first that all four parameters are jointly estimated by the proposed method. Fig. 1 displays experimental probability density functions (pdf) of estimates of each $\boldsymbol{\theta}_{\mathrm{RST}}$ component for TP #1. These pdfs are shown for the three estimators ($\mathrm{ML}_{\mathrm{fBm}}$, NGF, and MI) proved to be the best in our comparison. In addition, Gaussian pdfs $N(\boldsymbol{\theta}_{\mathrm{RST}}(i), \boldsymbol{\sigma}_{\mathrm{RST}}(i)^2)$ are shown as dashed curves for comparison with the distribution predicted by theory. Table 3 compares the $\mathrm{ML}_{\mathrm{fBm}}$, NGF, ML, NCC, and SSD estimators in terms of estimates bias. Fig. 2 presents data in terms of robust standard deviation $s_i$ just defined above.

The following observations can be drawn:

1. The mean percentage of outlying estimates roughly determined as $P\left(\left|\hat{\boldsymbol{\theta}}_{\mathrm{RST}}(i) - \boldsymbol{\theta}_{\mathrm{RST}}(i)\right| > 4s_i\right)$ is about 1% for the NGF estimator, 2.5% for the NCC and MI estimators, and 7% for the SSD estimator. For the $\mathrm{ML}_{\mathrm{fBm}}$ estimator, this value is only about 0.1%, i.e. the smallest.

2. The close proximity of experimental pdf for the $\mathrm{ML}_{\mathrm{fBm}}$ estimator with the Gaussian distribution can be clearly stressed (see pdfs in Fig.1). More in detail, according to Lilliefors goodness-of-fit test [36], the hypothesis of normality for $\Delta t$, $\Delta s$, $\alpha$ and $\Delta r$ estimate distributions can be accepted for the $ML_{fBm}$ estimator at significance level 5% for all TPs except for TP #2. The rest of estimators pass the normality test (after removing the abovementioned outliers) only for TP #5 (the MI method has also passed the normality test for TP #10; NGF method for TP #10 and #8).

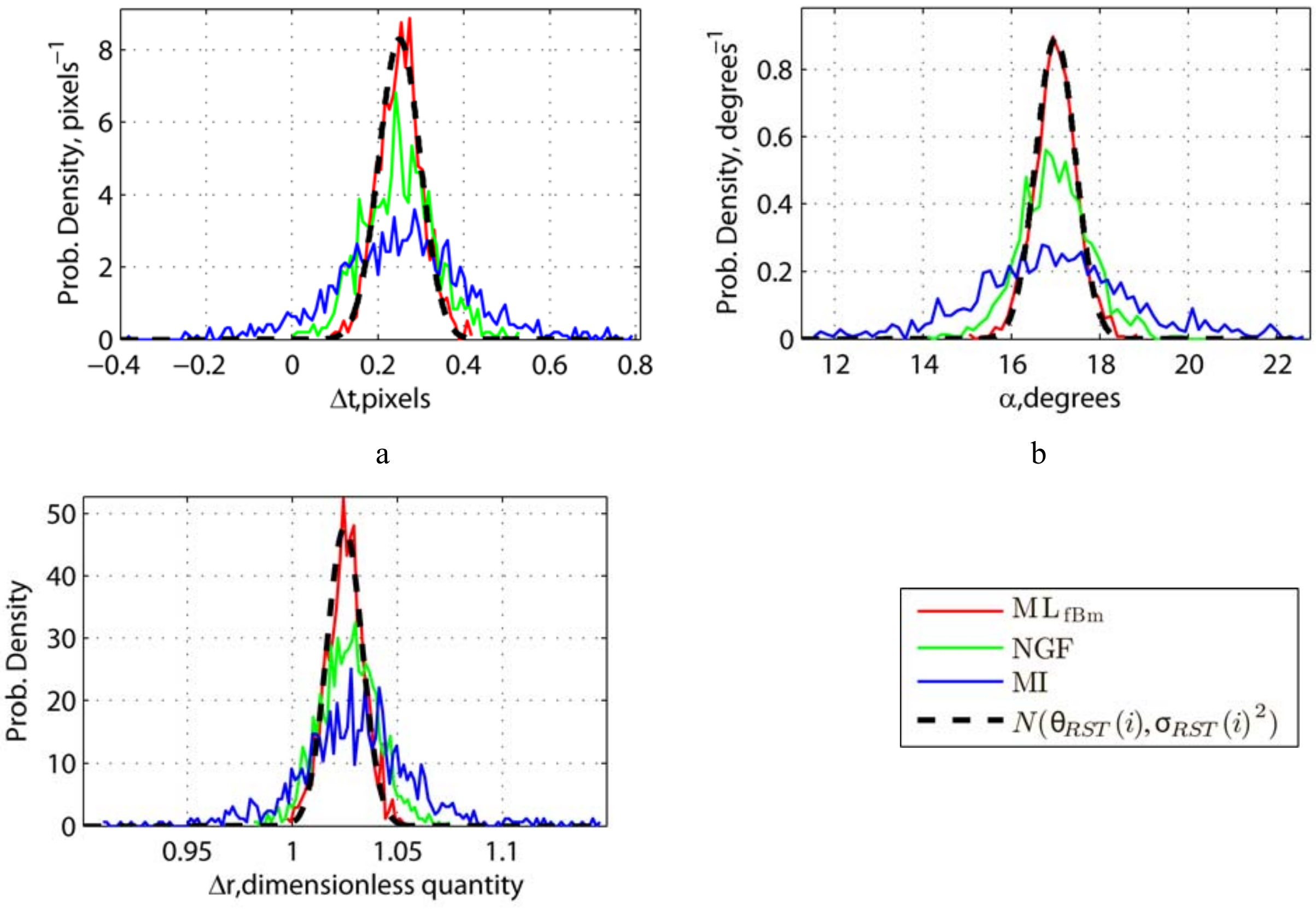

a b c

Fig. 1. Experimental pdfs of the RST parameter estimates for TP #1: horizontal translation (a), rotation angle (b) and scaling factor (c). The $ML_{fBm}$ data are shown as red curves, the NGF data - as green curves, the MI data - as blue curves and the theoretical pdfs $N\left(\boldsymbol{\theta}_{\mathrm{RST}}(i), \boldsymbol{\sigma}_{\mathrm{RST}}(i)^2\right)$ - as dashed black curves

3. For each estimator compared and each RST parameter, Table 3 shows minimum, maximum, MAD, and STD bias values obtained over all 10 TPs. The proposed $ML_{fBm}$ always shows the best results (ranked in the second position for MAD measure in only one case for translation $\Delta t$) in terms of bias maximum deviation interval (difference between max and min values), MAD and STD measures. For both NCC and SSD estimators, large errors are possible (they are responsible for increasing significantly the difference between max and min values and STD). We have found TP #2 (weak correlation between RI and TI) to be the worst case for efficacy of the MI, NCC and SSD estimators. In terms of MAD, the $ML_{fBm}$ reduces bias by a factor of about 2 for translation estimates, by 4…7 times for rotation angle and about 10 times for the scaling factor compared to the NGF, MI, NCC and SSD.

Table 3. Min, max, MAD and STD values of bias (multiplied by $10^3$) of translation (measured in pixels), of rotation angle (measured in degrees) and of scaling factor estimates obtained by the five methods

| Estimator | $\boldsymbol{\theta}_{RST}$ | min | max | MAD | STD | $\boldsymbol{\theta}_{RST}$ | min | max | MAD | STD |
|---|---|---|---|---|---|---|---|---|---|---|
| $ML_{fBm}$ | $\Delta t$, pixel | -8.5 | 2.9 | 2.8 | 3.9 | $\Delta s$, pixel | -8.8 | 3.7 | 0.8 | 3.2 |
| NGF | | -2.6 | 10.4 | 1.8 | 3.9 | | -19.4 | 6.1 | 2.8 | 7.3 |
| MI | | -3.5 | 22.8 | 4.2 | 9.2 | | -7.0 | 9.2 | 4.5 | 5.3 |
| NCC | | -32.0 | 4.4 | 5.8 | 10.8 | | -78.8 | 7.4 | 5.7 | 25.7 |
| SSD | | -46.0 | 2.3 | 9.1 | 13.9 | | -257.8 | 14.1 | 8.0 | 80.3 |
| $ML_{fBm}$ | $\alpha$, degree | -40.8 | 16.5 | 15.5 | 20.1 | $\Delta r$ | -1.6 | 0.4 | 0.3 | 0.6 |
| NGF | | -265.9 | 48.4 | 60.4 | 108.2 | | -0.9 | 27.5 | 3.0 | 8.4 |
| MI | | -178.2 | 171.2 | 114.0 | 126.3 | | 4.5 | 78.0 | 4.0 | 23.6 |
| NCC | | -927.1 | 62.3 | 64.4 | 294.4 | | -227.5 | 7.4 | 4.2 | 71.0 |
| SSD | | -1827.5 | 44.2 | 74.1 | 567.6 | | -246.7 | 13.8 | 3.5 | 76.3 |

4. In graphical form, estimation errors of the RST parameters are presented in Fig.2. For all estimators, intervals $\left[b_i - 3s_i, b_i + 3s_i\right]$ are shown as bars of specific colors. In addition, intervals $\left[-3\boldsymbol{\sigma}_{RST}(i), 3\boldsymbol{\sigma}_{RST}(i)\right]$ are given as semi-transparent bars. It is seen that according to $\left[b - 3s, b + 3s\right]$ intervals, the estimators can again be roughly ranked as follows: $ML_{fBm}$, NGF, MI, NCC, and SSD. In efficiency terms, the average efficiency (defined as $e = \frac{1}{4}\sum_{i=1}^{4} e_i$ ) of the proposed $ML_{fBm}$ estimator is about 90%, about 23% for the NGF estimator, 12-13% for the MI and NCC estimators, and, finally, about 6% for the SSD estimator. The behavior of the $ML_{fBm}$ estimator for TP #10 differs from the behavior observed for the rest of TPs and this will be discussed later in this Section. The NGF, MI and NCC estimators are less effective (by 20-50%) in estimating $\alpha$ and $\Delta r$ parameter as compared to translation parameters. TP #2 is the most challenging test point for all estimators, except the $ML_{fBm}$.

5. For TP #2, the average efficiency $e$ is about 85% for the $ML_{fBm}$, 3.5% for the NGF, 5.3% for the MI, 0.5% for the NCC and 0.25% for the SSD. This result is essential as TP #2 corresponds to the multitemporal and/or multimodal registration case (modeled by weak correlation between reference and template CFs). In this specific case and supported by experiment carried out on real data, the $ML_{fBm}$ estimator significantly outperforms even the MI method specially designed to cope with multimodal data.

6. For TP #10 (illustrating no geometrical transformation between reference and template CFs), the estimation error obtained with the $ML_{fBm}$ estimator is significantly lower than the value of the CRLB $\boldsymbol{\sigma}_{\mathrm{RST}}(i)$ for all components of the $\boldsymbol{\theta}_{\mathrm{RST}}$ vector (efficiency exceeds 100%). We mainly attribute this effect to a specific non-quadratic shape of log-LF (3) at this point.

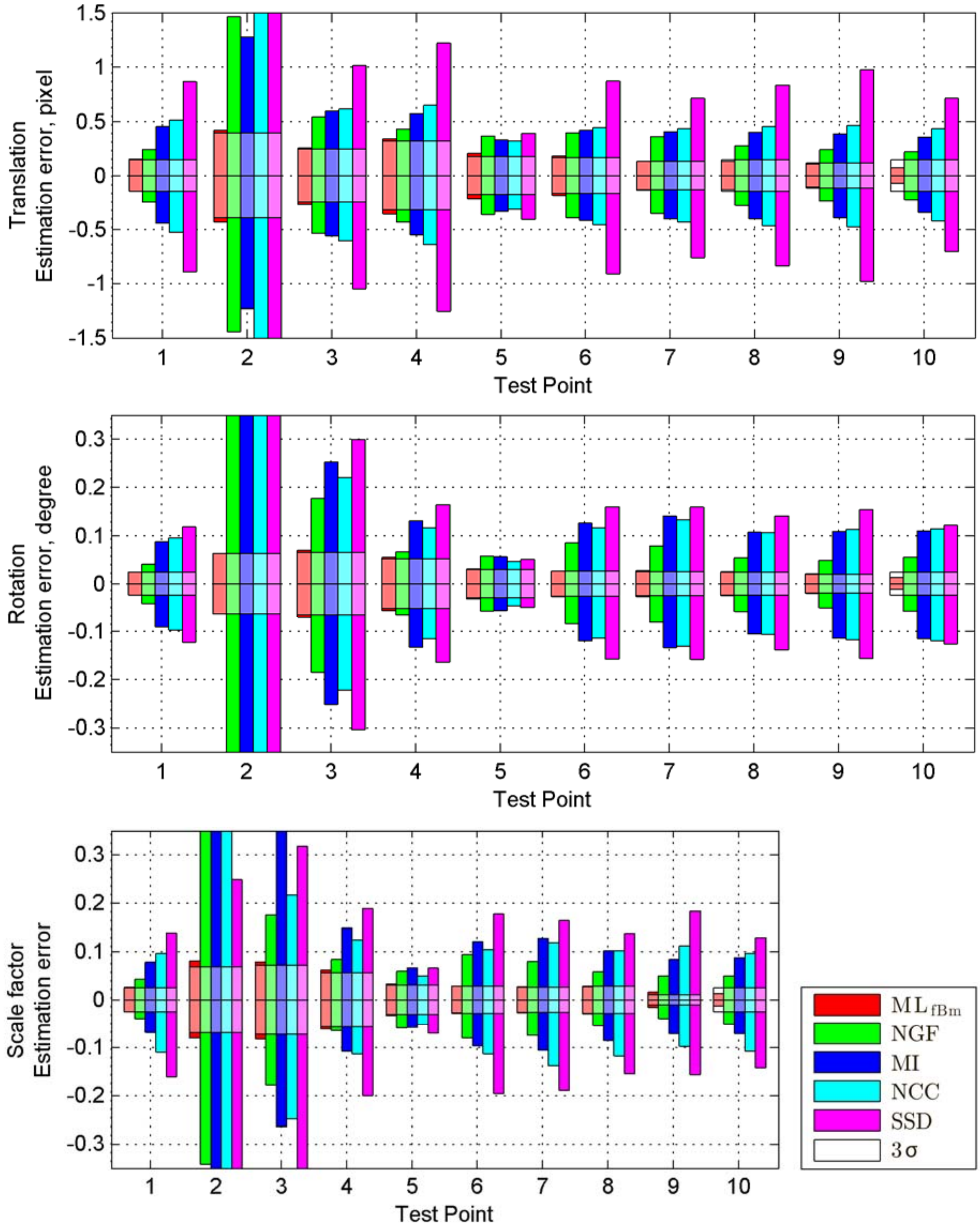

(a) (b) (c)

Fig. 2. Characteristics of RST parameters estimation errors for horizontal translation (a), rotation angle (b) and scaling factor (c) obtained by the five algorithms retained in the comparative study for all TPs #1-10

To better illustrate this, Fig. 3 displays a section of the mean log-LF w.r.t $\Delta r$ parameter and its approximation by the second-order Taylor expansion at the point $\boldsymbol{\theta}_{\mathrm{RST}} = (0,0,0,1)$. It is seen that the mean log-LF function decreases significantly faster than the quadratic function. As a result, the CRLB, which is based on second-order approximation of the log-LF shape, becomes inadequate. It underestimates the estimation accuracy of $\boldsymbol{\theta}_{\mathrm{RST}}$. We stress that this is only a local effect no longer visible on either side of (0,0,0,1). It clearly does not affect the performance of the $\mathrm{ML}_{\mathrm{fBm}}$ estimator, but it limits the adequacy of the derived CRLB at this particular point for samples of finite size.

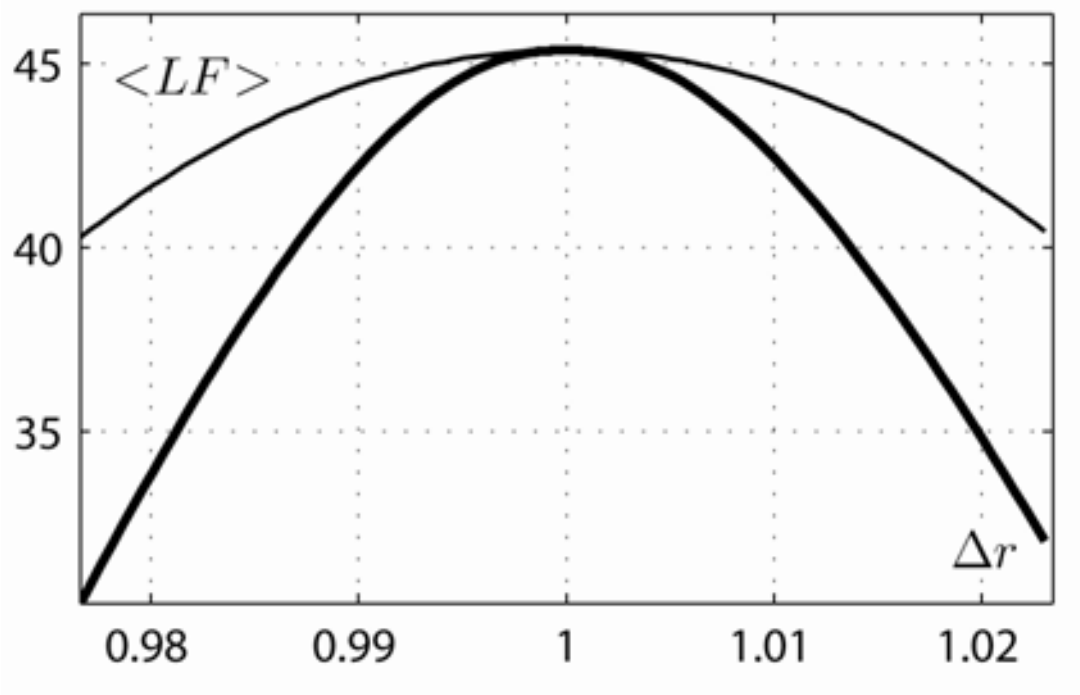

Fig. 3. Shape of log-LF (3) in the vicinity of the point $\boldsymbol{\theta}_{\mathrm{RST}} = (0,0,0,1)$ : the mean value of log-LF is shown as black thick curve, approximation by the second-order Taylor expansion - as black thin curve. Axis x spans the interval $[1-3\sigma_{\Delta r}, 1+3\sigma_{\Delta r}]$, where $\sigma_{\Delta r} = \boldsymbol{\sigma}_{\mathrm{RST}}(4)$ for TP #10.

### 3.3. Robustness to noise variance errors and complexity analysis of the $\mathrm{ML}_{\mathrm{fBm}}$

One more feature of the $\mathrm{ML}_{\mathrm{fBm}}$ estimator demands analysis: this is the only estimator involved in the comparison we have performed that directly requires knowledge of noise variance as an input. In practice, this value might be known with errors and the influence of these errors on the $\mathrm{ML}_{\mathrm{fBm}}$ performance should be investigated accordingly. Modern methods of blind noise variance estimation (including signal-dependent case) are known to perform well, with variance estimation error lying most of the time within the ±20% relative error interval (±10% for STD) [37]. So, we have performed additional experiments with setting erroneously both $\sigma_{n.\mathrm{RI}}, \sigma_{n.\mathrm{TI}}$ values with ±10% (and later ±20%) bias. Errors in noise variance lead to a limited increase of bias and estimation STD for all RST parameters. The most significant influence was seen at TP#4: $MSE(i)$ increased by about 5% (10%). For other TPs, the effect was significantly smaller, $MSE(i)$ increased by less than 4%. Therefore, the influence of noise variance estimation error on the performance of the $\mathrm{ML}_{\mathrm{fBm}}$ estimator can be reasonably neglected in practice.

Based on these results obtained on synthetic pure fBm data (with ground truth available), we can conclude that the proposed $ML_{fBm}$ estimator provides significant improvements compared to the four alternatives belonging to state-of-the-art. These improvements are seen in terms of standard deviation, bias and distribution shape of RST parameters estimates. However, we need to mention for sake of fairness that our estimator is significantly more computationally intensive as it requires operations with large sample correlation matrix. The cost of our current Matlab implementation of $ML_{fBm}$ estimator is 20s for estimation of RST parameter vector for one pair of CFs (reference CF is 23 by 23 pixels, template CF is 15 by 15 pixels) using Intel Core2 Duo T5450, 1.66 GHz. The similar operation with the same settings takes 0.6s for the NCC method (2D spline interpolation stage was found to be mainly responsible for the NCC method time cost), meaning that the $ML_{fBm}$ estimator is about 35 times slower than the NCC estimator. For larger sizes $N_{TI}$ and $N_{RI}$ of the CFs, this ratio will further increase. With this magnitude order, we have preferred to concentrate our efforts to demonstrate the $ML_{fBm}$ estimator potential for improving RST parameters estimation accuracy leaving efficient implementation for future work.

## 4. Performance analysis of the proposed estimator on real-life data

As a real-life example, we consider the registration of two images acquired by Hyperion and Landsat 8 sensors. Four among the five estimators considered previously completed by an extra one will be comparatively assessed on this pair of datasets. Thus, the comparison includes the $ML_{fBm}$, NCC, MI, NGF estimators and the LSM algorithm introduced in [8] at the fine registration stage. The latter algorithm is based on cross-correlation similarity measure and it is more suitable for real-life data than SSD.

### 4.1. Test data

Recall that Hyperion sensor [38] acquires hyperspectral images in 242 spectral bands with spectral resolution of about 10nm. Spectral range from 355.59 nm to 2577.08 nm is covered by two spectrometers (not all bands are active): VNIR (bands ## 1…70; 355.59… 1057.68 nm) and SWIR (bands ## 71…242; 851.92… 2577.08nm). Landsat 8 satellite [39] bears two pushbroom multispectral sensors, Optical Land Imager (OLI) and Thermal InfraRed Sensor (TIRS). OLI collects data from nine spectral bands (433…1390 nm; spatial resolution is 15/30 m), and TIRS acquires data in two spectral bands (10.30…12.50 μm; spatial resolution is 100m). The main parameters of the Hyperion and Landsat 8 datasets that were used in our experiment [40] are specified in Table 4.

Table 4. Characteristics of the Hyperion and Landsat 8 test datasets.

| Parameter | Hyperion | Landsat 8 (OLI) |
|---|---|---|
| **Dataset related information** | | |
| Scene ID | EO1H1800252002116110KZ | LC81770252014065LGN00 |
| Acquisition time | 26.04.2002 | 06.03.2014 |
| Path/ Row | 180/25 | 177/25 |
| Site Latitude/Longitude, degrees | 49.4339/32.0678 | 48.8497/31.6597 |
| Processing Level | L1R | L1T |
| Look angle, degrees | 9.7073 | 0 (nadir) |
| **Sensor related information** | | |
| Number of rows/columns | 3129/256 | 8061/7941 (reflective bands B1-B7)<br>16121/15881 (reflective band B8) |
| Spatial resolution, m | 30.38 | 30 (B1-B7, B9) or 15 (B8) |
| Swath, km | 7.7 | 185 |
| Orbit | Sun-synchronous; altitude is 705 km | Sun-synchronous; altitude is 708 km |

Among the 242 Hyperion bands, band #25 (VNIR; 599.80 nm) has been selected as the reference image. The Landsat 8 band B1 (OLI; 433…453 nm) is our template image. Spatial resolution of both bands is 30 m. We will later consider a more complex case when reference and template images have different spatial resolution. For this goal, we consider Landsat8 band B8 (OLI, 500…680 nm; panchromatic; spatial resolution is 15m) as template image.

Different acquisition settings (12 years difference in acquisition time, different wavelengths and spectral widths) make Landsat 8 to Hyperion registration a multitemporal or even a “mild” multimodal registration problem (true multimodality involves data acquired by sensors of different physical nature). This can be clearly seen from Fig. 4 that shows registered Hyperion (Fig. 4a) and Landsat 8 (Fig. 4b) bands. Different spatial resolutions complicate this problem even further.

To cope with the relief influence on Hyperion image, the fragment of ASTER Global Digital Elevation Map (GDEM) [40] covering the study area was used. DEM was manually registered to the Hyperion image (Fig. 4c). Relief for the study area is quite flat with elevation varying from 50 to 243 m (the mean elevation value is 113 m).

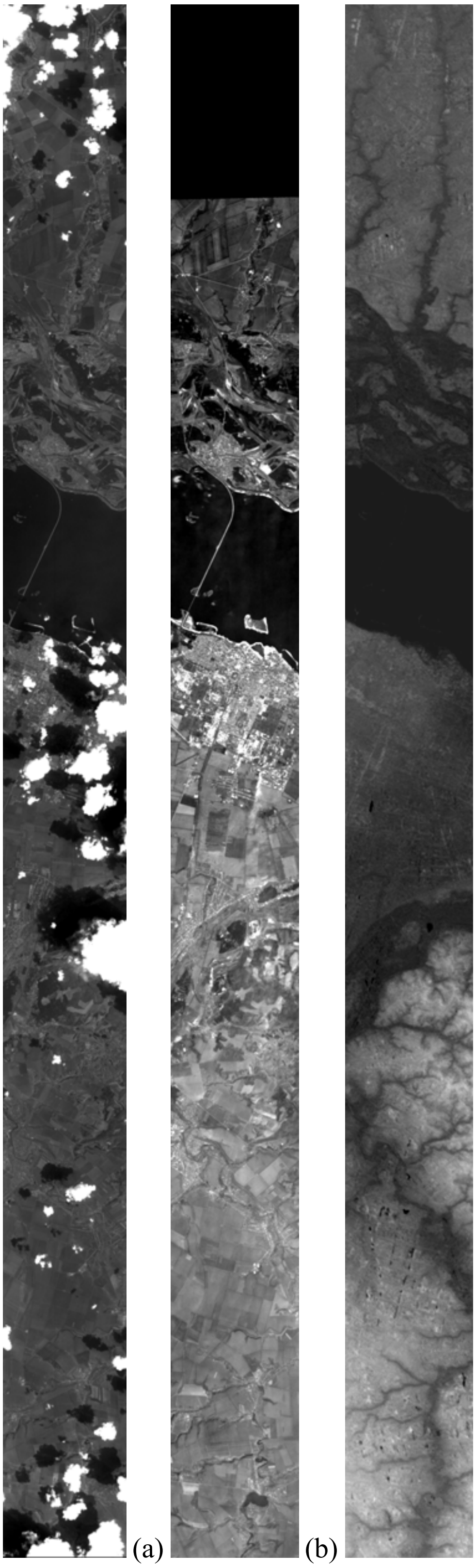

(a) (b) (c)

Fig. 4. Registered Hyperion band #25 (a), Landsat 8 band B1 (b) and DEM (c). Gray levels ranging from black to white cover intensity ranges 1100…3800 for Hyperion, 8500…9600 for Landsat 8 and 50…250m for DEM. Images size is 256 by 3129 pixels.

Relief influence in cross-track direction was systematically corrected at all stages described below based on Hyperion image acquisition parameters in Table 4. Lansat 8 image is terrain corrected, no additional correction is needed.

Noise parameters for the Hyperion and Landsat 8 datasets have been determined based on blind signal-dependent noise parameters estimation method [30] and according to the results obtained in [31]. Specifically, we have set the following noise model for both images:

$$\sigma_n^2 = \sigma_{n.SI}^2 + I\sigma_{n.SD}^2, \qquad (10)$$

where $I$ is the image intensity, $\sigma_{n.SI}^2$ and $\sigma_{n.SD}^2$ are the noise parameters that relate to signal-independent and signal-dependent components, respectively.

According to our estimates, $\sigma_{n.SI}$=8.3448 and $\sigma_{n.SD}$=0.2672 for the Hyperion band #25 and $\sigma_{n.SI}$=0 and $\sigma_{n.SD}$=0.1175 for the Landsat 8 band B1. When registering CFs of the two bands, noise variances $\sigma_{n.RI}^2$ and $\sigma_{n.TI}^2$ for each pair of reference/template CFs are obtained according to (10) by substituting $\sigma_{n.SI}$, $\sigma_{n.SD}$ with their estimates specified above and $I$ with CF mean intensity.

### 4.2. Coarse and fine registration stages

To register Landsat8 to Hyperion images, we adapted a two-stage approach that includes subsequent coarse and fine registration stages. At the coarse registration stage, we used the affine transformation model

$$\begin{pmatrix} i_{TI} \\ j_{TI} \end{pmatrix} = \mathbf{A}_{\text{HtoL}} \begin{pmatrix} i_{RI} \\ j_{RI} \end{pmatrix} + \mathbf{d}_{\text{HtoL}}, \quad (11)$$

where $(i_{RI}, j_{RI})$ denote terrain corrected row and column indices of the reference image, $(i_{TI}, j_{TI})$ denote row and column indices of the template image, $\mathbf{A}_{\text{HtoL}}$ is 2 by 2 matrix and $\mathbf{d}_{\text{HtoL}}$ is 2 by 1 translation vector, the lower subscript 'HtoL' means transformation from Hyperion to Landsat 8 image coordinate system.

Initially, Hyperion and Landsat 8 images were registered based on the corners longitude and latitude provided with each image. This registration occurred to be very inaccurate with errors up to 300 pixels in the along-track direction. To refine this result, we have applied automatic registration based on SURF descriptor [41] followed by RANSAC algorithm [42] to estimate affine transformation parameters in the presence of outliers. In this manner, registration error was reduced down to 2 pixels (this has been verified based on 15 manually selected control points).

Applying RQ-decomposition to $\mathbf{A}_{\text{HtoL}}$, we have found that the rotation angle and scaling factor between Hyperion and Landsat 8 were $\alpha_0 = 16.93^\circ$ and $\Delta r_0 = 1.0245$, respectively. The values $\alpha_0$ and $\Delta r_0$ have been later used as an initial guess of RST parameters at the fine registration stage.

The fine registration is next performed in three steps: 1) control fragments selection, 2) registration of each pair of CFs using one of the five estimators in comparison, 3) refinement of the affine transformation parameters $\mathbf{A}_{\text{HtoL}}$ and $\mathbf{d}_{\text{HtoL}}$.

### 4.3. CFs selection procedure

The CFs selection procedure includes the following stages:

1. The reference image is tiled by non-overlapping reference CFs of size $N_{\text{RI}} \times N_{\text{RI}}$ with coordinates $\left(i_{RI}(k), j_{RI}(k)\right)$, where $k$ denotes CF index (for notation simplicity, we will omit this index when some operation is applied to all CFs).

2. For each reference CF centered at $(i_{RI}, j_{RI})$ at the reference image RI, the corresponding position $(i_{TI}, j_{TI})$ of template CF at the template image TI is calculated using (11). As $i_{TI}$ and $j_{TI}$ can be fractional numbers, we center template CF position at $([i_{TI}],[j_{TI}])$, where $[\cdot]$ is the operation of rounding to the nearest integer. For each CF, $\boldsymbol{\theta}_{\mathrm{RST}}$ is initialized as $\boldsymbol{\theta}_{\mathrm{RST.IG}} = (\Delta t_0, \Delta s_0, \alpha_0, \Delta r_0)$, where $\Delta t_0 = i_{TI} - [i_{TI}]$ and $\Delta s_0 = j_{TI} - [j_{TI}]$ are initial subpixel translations.
3. All CFs are grouped into four groups according to two attributes: Normal vs. not Normal and isotropic vs. anisotropic texture. Group I is for Normal and Isotropic textures, group II is used for Normal but Anisotropic, group III - for Isotropic but not Normal, and group IV - for both not Normal and Anisotropic. The reason for such grouping is that texture anisotropy and abnormality does not match with fBm model. By preclassifying CFs, we seek to evaluate the robustness of the $\mathrm{ML_{fBm}}$ estimator to texture deviations from fBm model. From the four groups I…IV, group I contains CFs that best match the fBm approach.

Anisotropic textures have been detected by calculating autocorrelation function of template CF, $r(\Delta i, \Delta j)$, approximating it by second order polynomial $r(\Delta i, \Delta j) = a\Delta i^2 + b\Delta j^2 + 2c\Delta i\Delta j + d\Delta i + e\Delta j + f$ and calculating eigenvalues $\lambda_{\max}$ and $\lambda_{\min}$ of the matrix $\begin{bmatrix} a & c \\ c & b \end{bmatrix}$. A pair of reference/template CFs is considered isotropic if $\lambda_{\max} / \lambda_{\min} < 2$, otherwise this pair is considered as anisotropic. A pair of reference/template CFs is considered Normal if both vertical and horizontal increments of template CF with unity lag pass the Lilliefors normality test [36] with significance level 1%. In total, 1500 pairs of CFs have been detected suitable for our registration processing scheme, among them 416 belong to group I, 138 to group II, 473 to group III, and 473 to group IV.

### 4.4. Ensuring global convergence

Initializing the $ML_{fBm}$ estimator by the vector $\boldsymbol{\theta}_{RST.IG}$ previously defined (in item 2 of subsection 4.3) does not, in general, assure convergence to the global maximum. Indeed, the magnitude of the coarse registration error with respect to translation parameter is about 2 pixels. With this, it has been experimentally found that the attracting area of the global maximum of the proposed log-LF with respect to translation is about ±0.6 pixel wide. This clearly means that $\boldsymbol{\theta}_{RST.IG}$ could be outside the attracting area of the global log-LF maximum leading to erroneous estimates. To assure global convergence, we have considered the so called multi-start optimization technique with nine different initial guesses for $\boldsymbol{\theta}_{RST.IG}$ : $\boldsymbol{\theta}_{RST.IG} = \left(\Delta t_0 + \Delta t_{shift}, \Delta s_0 + \Delta s_{shift}, \alpha_0, \Delta r_0\right)$ , where $\Delta t_{shift}, \Delta s_{shift} = -1, 0, 1$. Convergence of the $ML_{fBm}$ is illustrated in Fig. 5 where nine convergence paths are superimposed on the 2D cross-section of the log-LF: each point $\left(\Delta t, \Delta s\right)$ corresponds to the maximum log-LF value with respect to $(\sigma_{x.RI}, \sigma_{x.TI}, H, k_{RT})$ vector, setting the two remaining parameters as $\alpha = \alpha_0$, $\Delta r = \Delta r_0$.

Fig. 5a shows a typical convergence scenario, seen for majority of CFs. The initial guess $\boldsymbol{\theta}_{RST.IG} = \left(\Delta t_0, \Delta s_0, \alpha_0, \Delta r_0\right)$ lies within the main lobe of the log-LF and leads to correct final estimation result. An example of an opposite situation when $\left(\Delta t_0, \Delta s_0, \alpha_0, \Delta r_0\right)$ does not belong to the main lobe of the log-LF is shown in Fig. 5b. In this case, convergence to the global maximum is truly assured by other initial guesses.

The same procedure has been used indifferently for the four NGF, MI, NCC, and LSM estimators: the corresponding similarity measures are thus minimized nine times starting each time from a different initial guess among the nine considered. The estimate that corresponds to the absolute minimum of each similarity measure is just taken as the final estimate.

For each pair of CFs, we have obtained five estimates $\hat{\boldsymbol{\theta}}_{RST.estimator} = \left(\widehat{\Delta t}, \widehat{\Delta s}, \hat{\alpha}, \widehat{\Delta r}\right)$, where the subscript "estimator" takes one or the other of the values "MLfBm", "NGF", "MI", "NCC" or “LSM”. Similarly to Section 3, we set $N_{\mathrm{RI}} = 23$ and $N_{\mathrm{RI}} = 15$ for all estimators. For the MLfBm estimator, additional results are obtained as auxiliary data: texture parameter vector $\hat{\boldsymbol{\theta}}_{\mathrm{texture}} = \left(\hat{\sigma}_{x.\mathrm{RI}}, \hat{\sigma}_{x.\mathrm{TI}}, \hat{H}, \hat{k}_{\mathrm{RT}}\right)$ and estimate $\hat{\boldsymbol{\sigma}}_{\mathrm{RST}}$ of the $\boldsymbol{\sigma}_{\mathrm{RST}}$ vector. The latter is simply found by substituting $\hat{\boldsymbol{\theta}}_{\mathrm{texture}}$ and $\hat{\boldsymbol{\theta}}_{\mathrm{RST.ML_{fBm}}}$ into (6).

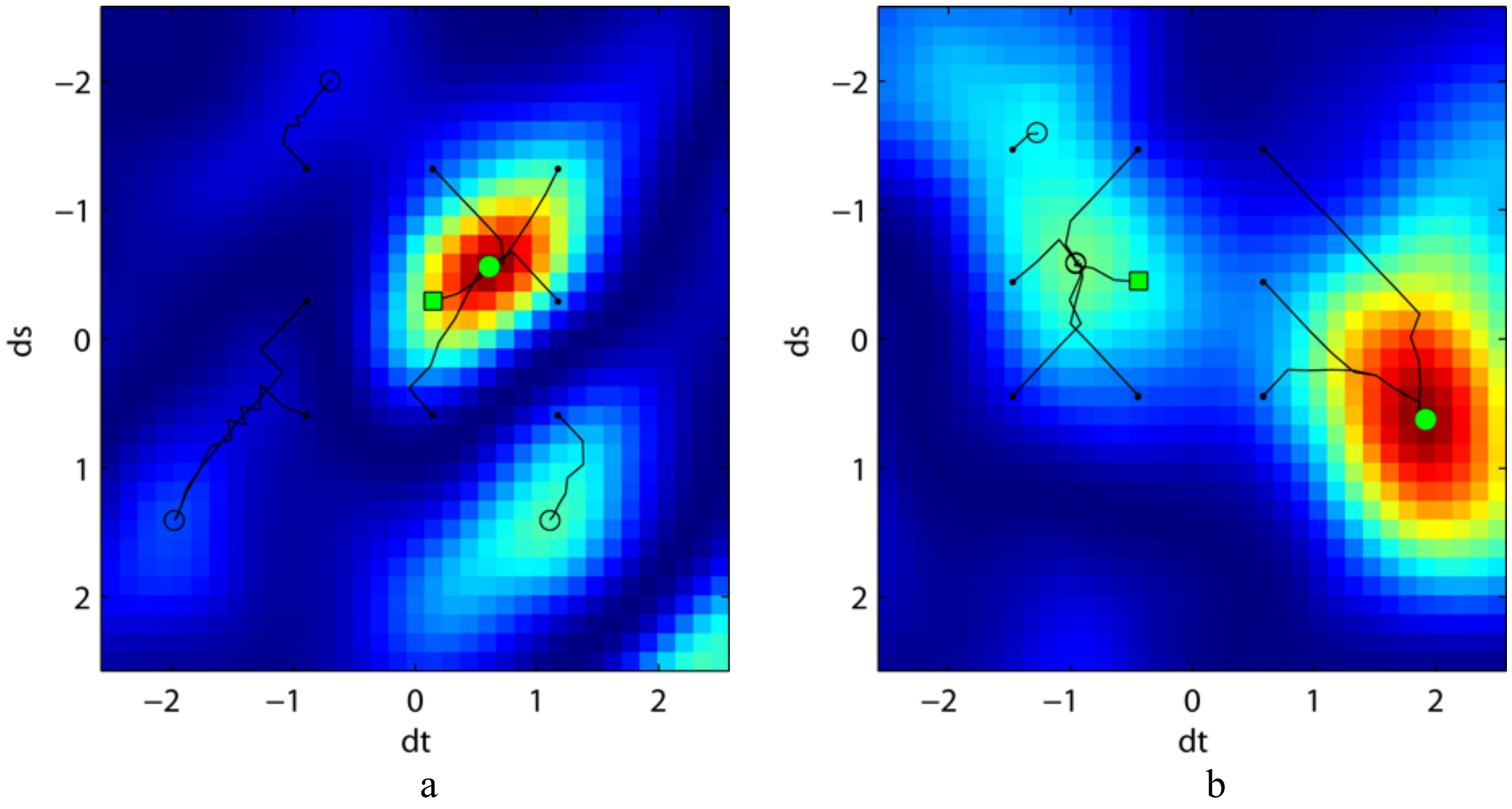

Fig. 5. Convergence of the $\mathrm{ML_{fBm}}$ estimator for two CFs. Larger log-LF values are shown in red color, lower – in blue color. Nine convergence paths are shown in black; the starting points are marked as “•”, the end points - as “o”; the point $\left(\Delta t_0, \Delta s_0\right)$ - as green “□”marker; the global log-LF maximum - as green “o” marker. Initial guess is within (a) and outside (b) the mail lobe of the log-LF.

Due to lack of ground truth, all estimates $\hat{\boldsymbol{\theta}}_{\mathrm{RST.estimator}}$ will be compared with the output of the subsequent fine registration stage $\hat{\boldsymbol{\theta}}_{\mathrm{RST.fine}}$ (as explained at the end of subsection 4.2). At this stage, we used RANSAC algorithm fed with the $\mathrm{ML_{fBm}}$ estimates for CFs of group I to get refined estimates of the global affine transform $\mathbf{A}_{\mathrm{HtoL.fine}}$ and $\mathbf{d}_{\mathrm{HtoL.fine}}$.

### 4.5. Test of hypothesis of identical Hurst exponent values for Hyperion and Landsat 8 images

Before analyzing quantitatively the accuracy of RST parameters estimation, let us check validity of the hypothesis stating that the same Hurst exponent can be used for reference and template CFs. Recall that this hypothesis has been accepted above to derive the correlation matrix (2). To this end, for each pair of CFs, two estimates of the Hurst exponent, $\hat{H}_{\mathrm{RI}}$ and $\hat{H}_{\mathrm{TI}}$, were obtained independently for reference and template CFs. We can observe from distribution of the pairs of estimates ($\hat{H}_{\mathrm{RI}}, \hat{H}_{\mathrm{TI}}$) (see Fig. 6) that they are concentrated enough along the line $H_{\mathrm{RI}} = H_{\mathrm{TI}}$. Correlation between $\hat{H}_{\mathrm{RI}}$ and $\hat{H}_{\mathrm{TI}}$ is 0.55. Thus, the hypothesis $H_{\mathrm{RI}} = H_{\mathrm{TI}}$ can be reasonably accepted.

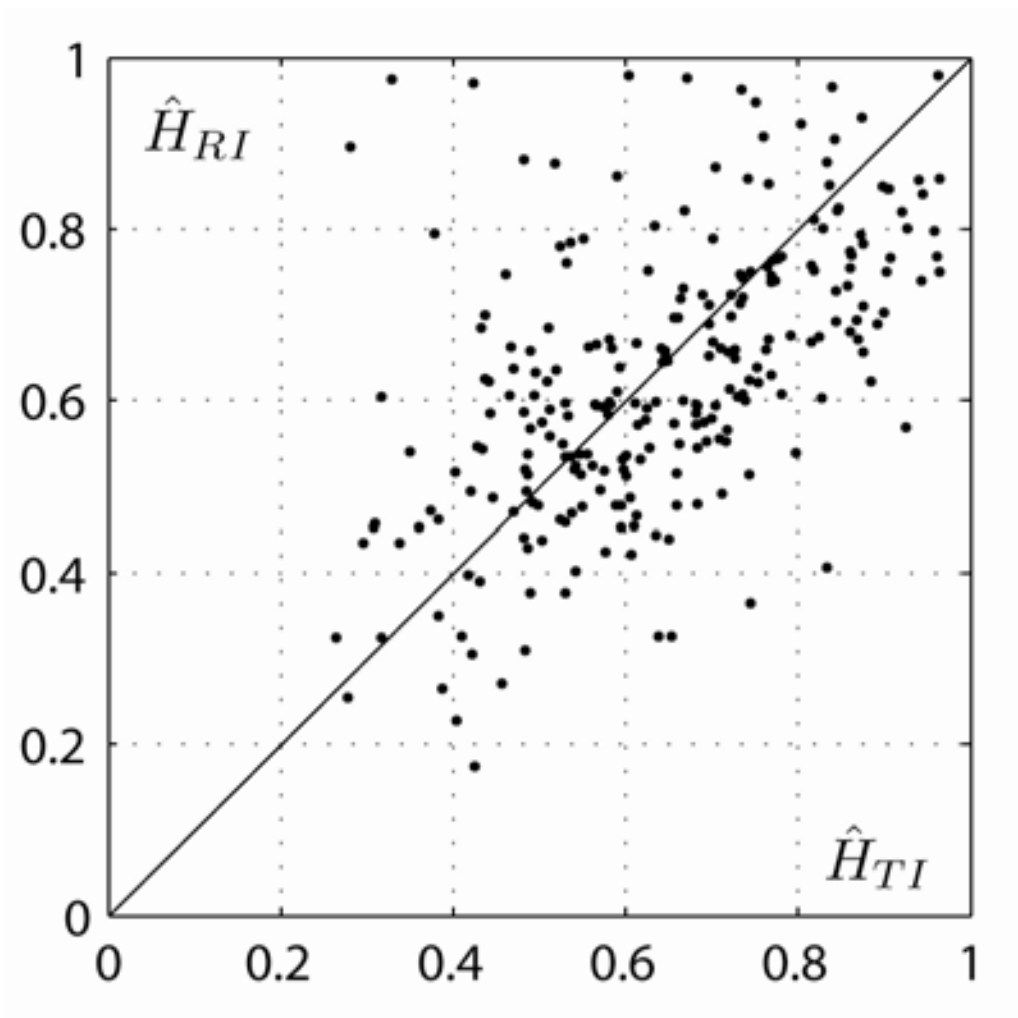

Fig. 6. Distribution of pairs ($\hat{H}_{\mathrm{RI}}, \hat{H}_{\mathrm{TI}}$) for CFs of group I

### 4.6. Quantitative analysis measures

Let us now analyze the estimation accuracy of $\boldsymbol{\theta}_{\mathrm{RST}}$ vector for different CFs. The following three measures are adopted for this purpose: probability of outlying estimates, absolute error STD and normalized error STD. Below, we introduce and briefly discuss each measure.

Typically, an outlying estimate is defined as an estimate lying outside a circle with a predefined radius centered at the true value of parameters vector. For translation parameter estimates, a typical value of this radius is one pixel [43]. This definition is intuitively clear but subjective by nature.

Indeed, it is clear that different pairs of CFs can be suitable for registration in a different degree. For example, a higher value of $\sigma_{x.\mathrm{RI}} / \sigma_{n.\mathrm{RI}}$ and $\sigma_{x.\mathrm{TI}} / \sigma_{n.\mathrm{TI}}$ ratios (that are related to SNR measure) and a higher magnitude of correlation coefficient $k_{\mathrm{RT}}$ should lead to a more accurate registration. Within the proposed approach, this variability can be characterized by the corresponding CRLBs (elements $\sigma_{\Delta t}$, $\sigma_{\Delta s}$, $\sigma_{\alpha}$ and $\sigma_{\Delta r}$ of vector $\hat{\boldsymbol{\sigma}}_{\mathrm{RST}}$). For the considered registration scenario, $\hat{\sigma}_{\Delta t}$ and $\hat{\sigma}_{\Delta s}$ vary from 0.025 to 2 pixels with the mode located at ≈0.1 pixel; $\hat{\sigma}_{\alpha}$ varies from 0.2 to 15 degrees with the mode about 0.9 degrees; $\hat{\sigma}_{\Delta r}$ varies from 0.004 to 0.3 with the mode around 0.018. Overall, for all components, the standard deviation of estimation error can exhibit a 75-fold variation. This quite high variation indicates that it is impossible to detect outlying estimates by applying the same threshold to all pairs of CFs.

However, an outlying estimate can be more properly defined if a reasonable distribution of normal estimates is assumed. This can be done using $\mathbf{C}_{\boldsymbol{\theta}_{\text{RST}}}$ bound. Recall that asymptotic distribution of $\boldsymbol{\theta}_{\text{RST}}$ vector estimate by an efficient unbiased estimator is $N(\boldsymbol{\theta}_{\text{RST0}}, \mathbf{C}_{\boldsymbol{\theta}_{\text{RST}}})$, where $\boldsymbol{\theta}_{\text{RST0}}$ denotes the true RST parameter vector. Here, we use the estimate $\hat{\boldsymbol{\theta}}_{\text{RST.fine}}$ as $\boldsymbol{\theta}_{\text{RST0}}$. For a practical estimator, $\boldsymbol{\theta}_{\text{RST}}$ estimates distribution should be more or less close to $N(\boldsymbol{\theta}_{\text{RST0}}, \mathbf{C}_{\boldsymbol{\theta}_{\text{RST}}})$. Thus, to detect outliers, we need to test zero hypothesis that $\hat{\boldsymbol{\theta}}_{\text{RST}}$ follows $N(\boldsymbol{\theta}_{\text{RST0}}, \mathbf{C}_{\boldsymbol{\theta}_{\text{RST}}})$ distribution against alternative hypothesis that $\hat{\boldsymbol{\theta}}_{\text{RST}}$ does not obey $N(\boldsymbol{\theta}_{\text{RST0}}, \mathbf{C}_{\boldsymbol{\theta}_{\text{RST}}})$. The sufficient statistics for this test is the quadratic form $Q = (\hat{\boldsymbol{\theta}}_{\text{RST}} - \boldsymbol{\theta}_{\text{RST0}})^T \mathbf{C}_{\boldsymbol{\theta}_{\text{RST}}}^{-1} (\hat{\boldsymbol{\theta}}_{\text{RST}} - \boldsymbol{\theta}_{\text{RST0}})$. We define accordingly an outlying estimate by the following rule:

$$Q > Q_{th}, \tag{12}$$

where $Q_{th}$ is a threshold. For the zero hypothesis, $Q$ should follow a $\chi^2$ distribution with four degrees of freedom (the number of RST parameters). At significance level $\alpha = 1 - 10^{-6}$ for $\chi^2(4)$ distribution, we get $Q_{th} = 33.3768$. Probability of outlying estimates can now be obtained as $P_{out} = P(Q > Q_{th})$.

Normalized errors vector is obtained by dividing each element of the absolute error $\Delta\boldsymbol{\theta}_{\text{RST}} = \hat{\boldsymbol{\theta}}_{\text{RST}} - \boldsymbol{\theta}_{\text{RST0}}$ by the corresponding element of $\boldsymbol{\sigma}_{\text{RST}}$ (potential STD value): $\delta\boldsymbol{\theta}_{\text{RST}} = \Delta\boldsymbol{\theta}_{\text{RST}}./\boldsymbol{\sigma}_{\text{RST}}$, where ./ defines pointwise matrix division. Below, we deal with standard deviation of absolute ($s_{\text{abs}.i}$) and normalized ($s_{\text{norm}.i}$) errors. These standard deviations are defined as in the Section 3 through MAD measure to prevent outliers influence.

### 4.7. Absolute errors analysis

Let us start with the analysis of absolute errors. For the CFs belonging to group I, the experimental pdfs of absolute errors corresponding to the $\text{ML}_{\text{fBm}}$ and MI estimators are shown in Fig. 7 (the NGF and NCC methods produced results similar to the MI). Pdfs were computed using kernel smoothing density estimate implemented in ksdensity Matlab function. It is seen that for the $\text{ML}_{\text{fBm}}$ estimator, these errors are characterized by the lowest variance and the absence of heavy-tails caused by outliers (two spikes in the scaling factor pdf for the MI method are due to constraints in the form of lower and upper bounds imposed on $\Delta r$ value).

Absolute error STDs $s_{\text{abs}.i}$ are given in Table 5. The general observation is that the $\text{ML}_{\text{fBm}}$ estimator offers substantial performance improvement over the NGF, MI, NCC, and LSM methods for all groups of CFs and all RST parameters: $s_{\text{abs}.i}$ decreases by 1.5…2.6 times for all RST parameters. The NGF, MI and NCC methods show similar performance for groups I-III. For group IV, the NGF outperforms the MI and NCC methods. For all groups, the LSM demonstrates the worst estimation accuracy.

All methods involved in the comparison carried out show similar performance for groups I-II, decreased performance for group III and even more significant decrease for group IV. Therefore, it can be concluded that texture anisotropy affects all five registration methods in a negative manner but only slightly, texture non-normality affects them more significantly and combination of these two factors degrades estimation even more significantly.

**4.8. Normalized error analysis**

The experimental pdfs of the quadratic form $Q$ defined in (12) for the $\text{ML}_{\text{fBm}}$ and MI estimators and calculated for CFs of group I are given in Fig. 8 (again the results for NGF and NCC are similar to MI). The threshold $Q_{th}$ is shown as the vertical black thick line. The pdf of statistic $Q$ for the $\text{ML}_{\text{fBm}}$ method is significantly more concentrated towards zero values and has smaller right-hand tail as compared to the one for the MI method. In quantitative sense, this leads to decreased normalized error STD and decreased percentage of outliers.

For group I, the percentage of outliers is only 10% for the $\text{ML}_{\text{fBm}}$ method but it increases up to 48% for the NGF, MI, and NCC methods. For groups II-IV, we see the same tendency as for absolute errors: the percentage of outliers slightly increases for groups II and III. This increase becomes significant for group IV. For the LMS method, the estimation errors are very significant. Due to this, almost all estimates are classified as outliers.

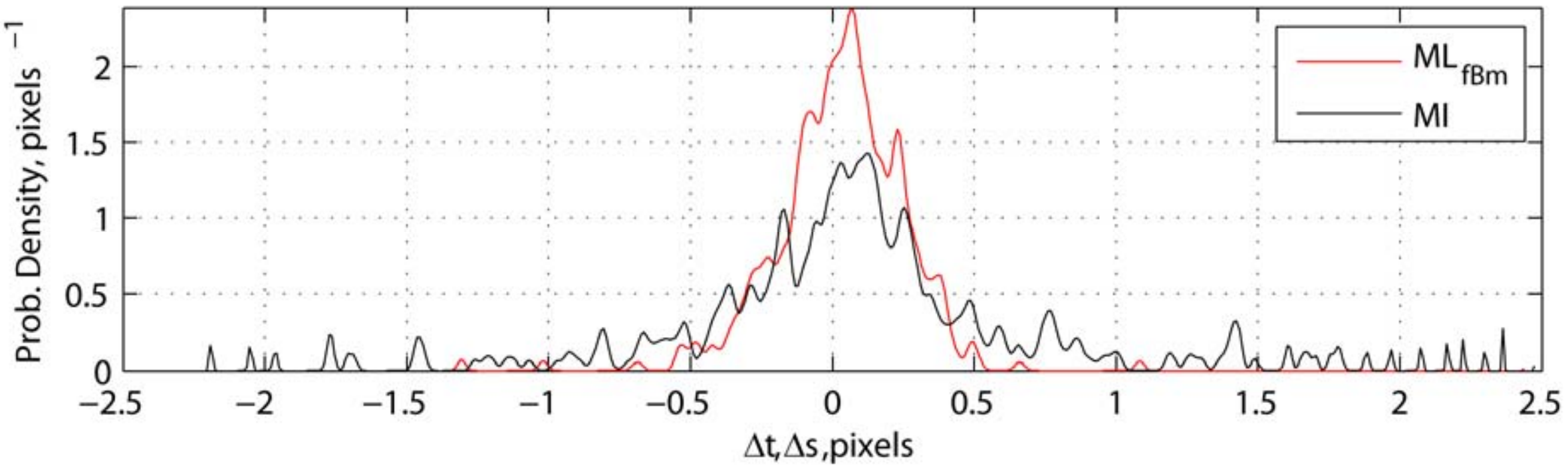

(a)

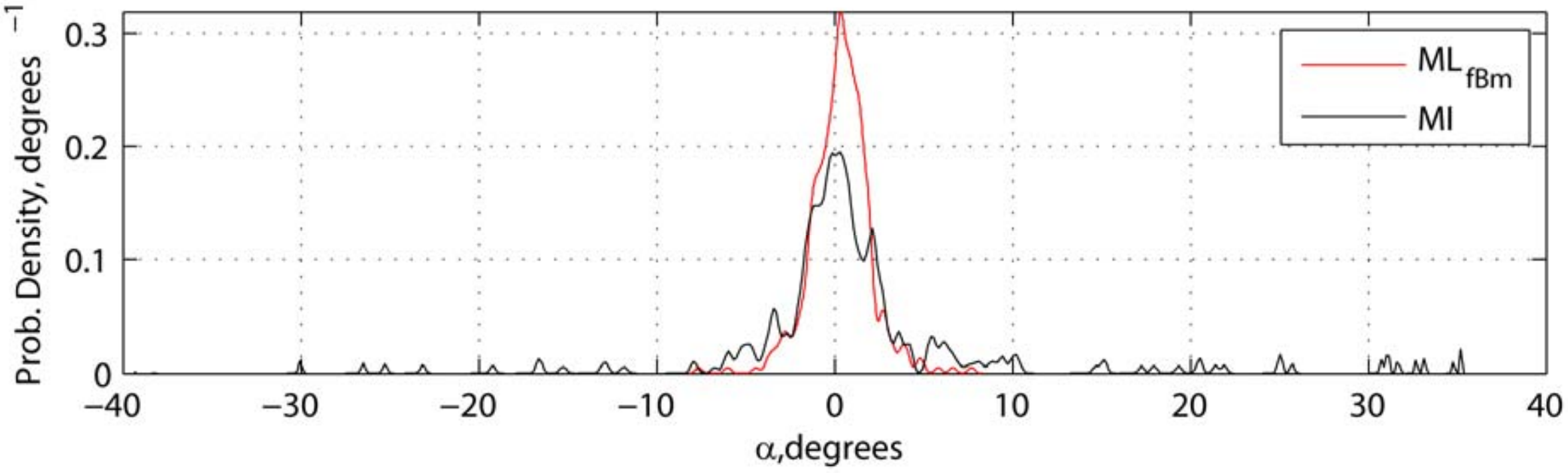

(b)

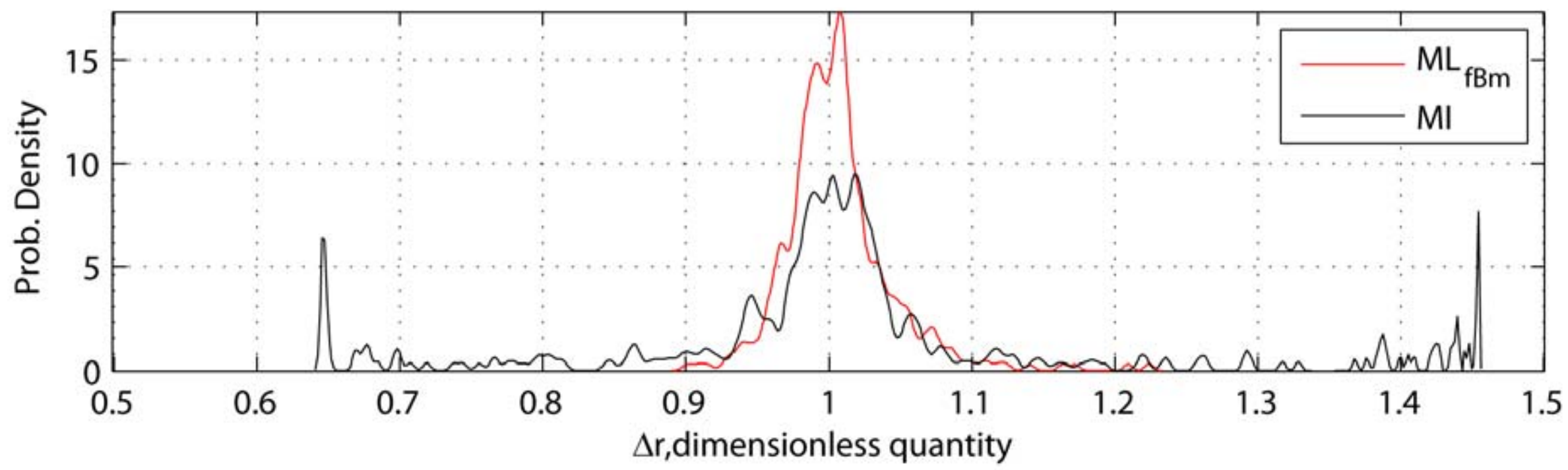

(c)

Fig. 7. Experimental pdfs of RST parameters estimates by the $ML_{fBm}$ and MI estimators.

a) translations, b) rotation angle and c) scaling factor. All pdfs are obtained for CFs from group I.

Table 5. Absolute errors analysis (the smallest STD values of absolute errors are shown in bold font)

| RST parameter | Estimator | Standard deviation of absolute errors (1.48MAD) | | | |
|---|---|---|---|---|---|
| | | Group I | Group II | Group III | Group IV |
| $\Delta t$ and $\Delta s$ | $ML_{fBm}$ | **0.198** | **0.229** | **0.226** | **0.309** |
| | NGF | 0.36 | 0.368 | 0.365 | 0.454 |
| | MI | 0.336 | 0.402 | 0.362 | 0.51 |
| | NCC | 0.311 | 0.355 | 0.378 | 0.651 |
| | LSM | 1.797 | 1.898 | 1.866 | 3.590 |
| $\alpha$ | $ML_{fBm}$ | **0.023** | **0.024** | **0.029** | **0.024** |
| | NGF | 0.044 | 0.052 | 0.046 | 0.043 |
| | MI | 0.046 | 0.031 | 0.047 | 0.044 |
| | NCC | 0.040 | 0.036 | 0.049 | 0.045 |
| | LSM | 0.181 | 0.173 | 0.216 | 0.268 |
| $\Delta r$ | $ML_{fBm}$ | **0.027** | **0.027** | **0.034** | **0.044** |
| | NGF | 0.054 | 0.050 | 0.055 | 0.079 |
| | MI | 0.049 | 0.059 | 0.068 | 0.088 |
| | NCC | 0.070 | 0.061 | 0.078 | 0.127 |
| | LSM | 0.177 | 0.189 | 0.206 | 0.345 |

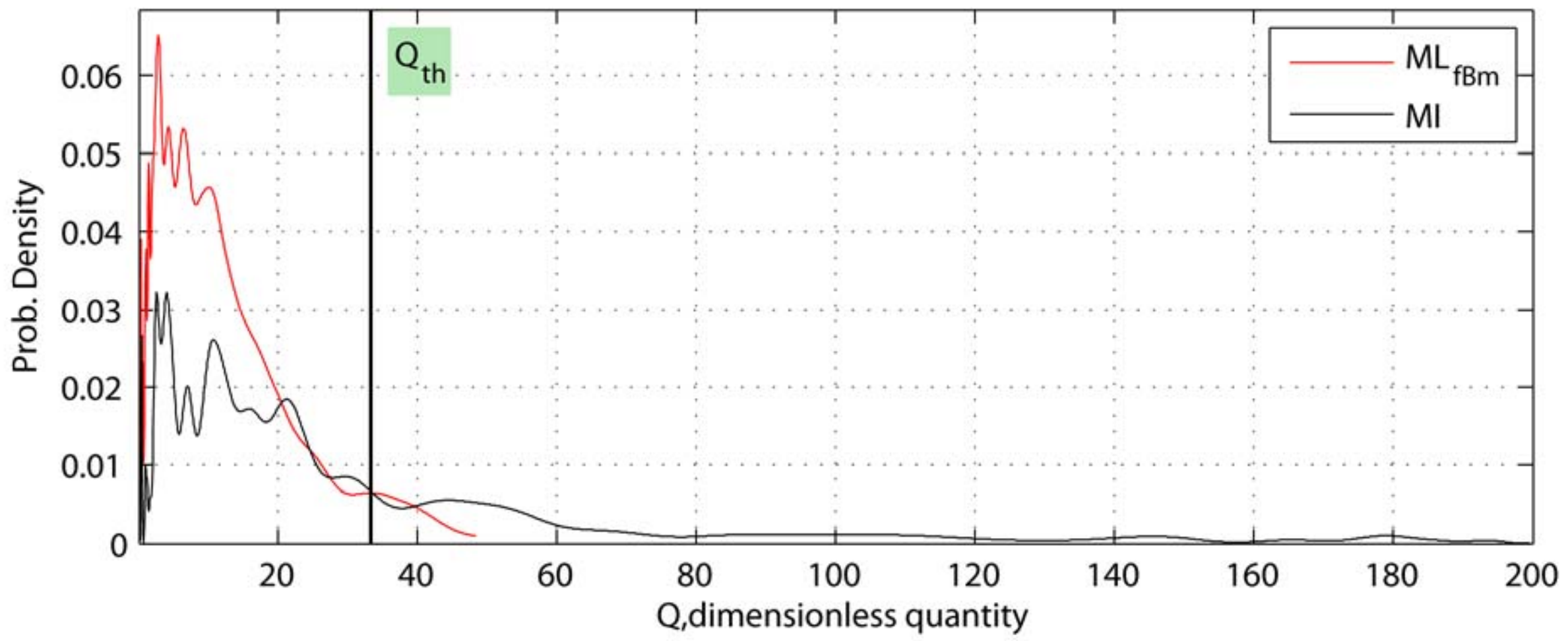

Fig. 8. Experimental pdf of the quadratic form $Q$ obtained for CFs from group I.

Table 6. Probability of outlying estimates of RST parameters, $P_{\text{out}}$, %

(the smallest probabilities of outlying estimates are shown in bold font)

| Estimator | Group I | Group II | Group III | Group IV |
|---|---|---|---|---|
| $\text{ML}_{\text{fBm}}$ | **10.10** | **20.29** | **23.89** | **44.82** |
| NGF | 48.56 | 55.07 | 56.03 | 66.81 |
| MI | 47.36 | 56.52 | 57.51 | 72.09 |
| NCC | 47.12 | 55.07 | 60.68 | 77.59 |
| LSM | 94.95 | 97.82 | 95.56 | 97.04 |

For a more detailed analysis, Table 7 summarizes standard deviations of the normalized errors $s_{\text{norm}.i}$ separately for each group of CFs and for each RST parameter. In addition, for group I the results are given in parenthesis for three subintervals of $k_{RT}$ values: less than 0.6, from 0.6 to 0.8 and larger than 0.8. The experimental results with different spatial resolutions between reference and template images are presented in the last column.

To better interpret data in Table 7, recall that for an efficient estimator and an accurate lower bound used for normalization, $s_{\text{norm}.i}$ should be close to unity. For CFs of group I, the $\text{ML}_{\text{fBm}}$ is very close to this ideal case with $s_{\text{norm}}$ about 2 for translation parameters and 1.55 for rotation angle and scaling factor. This corresponds to an efficiency of the $\text{ML}_{\text{fBm}}$ estimator w.r.t. $\mathbf{C}_{\mathbf{\theta}}$ bound of about 25…42% on real data whilst it is of 90% for the simulated pure fBm-samples. To our opinion, this is a reasonable price to pay for applying a model-based estimator to a complex pair of real datasets.

For the NGF, MI and NCC methods, $s_{\text{norm}}$ increases by a factor of 1.75…2 as compared to the $\text{ML}_{\text{fBm}}$ estimator. This observation remains for groups II-IV with the same tendencies as previously:

$s_{\text{norm}}$ increases slightly for all estimators in comparison for groups II and III, this increase becomes significant for group IV. For the LSM method, $s_{\text{norm}}$ is significantly larger than for other methods in comparison.

Table 7. Normalized errors analysis (the smallest STD values of normalized errors are shown in bold font)

| RST parameter | Estimator | Standard deviation of normalized errors (1.48MAD), dimensionless quantity | | | | |
|---|---|---|---|---|---|---|
| | | Group I all CFs (0.4…0.6/0.6…0.8/0.8…0.95) | Group II | Group III | Group IV | Group I (Landsat8 B8) |
| $\Delta t$ and $\Delta s$ | $ML_{fBm}$ | **2.02 (1.71/ 2.26/ 2.03)** | **2.5** | **2.44** | **3.67** | **2.67** |
| | NGF | 3.59 (4.26/ 3.35/ 3.31) | 4.35 | 4.01 | 5.43 | --- |
| | MI | 3.55 (4.11/ 3.62/ 2.49) | 4.28 | 3.96 | 5.74 | --- |
| | NCC | 3.37 (4.39/ 2.94/ 2.63) | 4.23 | 4.37 | 7.83 | 3.80 |
| | LSM | 17.45 (17.91/ 16.47/ 16.78) | 20.57 | 18.84 | 41.80 | --- |
| $\alpha$ | $ML_{fBm}$ | **1.54 (1.28/1.60/** 2.29) | **1.59** | **1.86** | **1.97** | **2.03** |
| | NGF | 2.9 (3.73/2.57/ 2.51) | 3.36 | 3.26 | 3.53 | --- |
| | MI | 2.85 (3.41/ 2.50/ **2.27**) | 2.23 | 3.36 | 3.51 | --- |
| | NCC | 2.58 (4.37/ 2.09/ 2.53) | 2.64 | 3.37 | 3.45 | 2.53 |
| | LSM | 11.19 (12.17/ 9.34/ 9.71) | 12.11 | 14.81 | 20.04 | --- |
| $\Delta r$ | $ML_{fBm}$ | **1.56 (1.59/ 1.39/** 1.78) | **1.84** | **2.08** | **3.06** | **2.24** |
| | NGF | 3.08 (3.91/ 3.04/ **1.76**) | 3.77 | 3.66 | 5.2 | --- |
| | MI | 3.15 (4.06/ 2.55/ 2.31) | 3.51 | 4.4 | 6.37 | --- |
| | NCC | 4.01 (7.39/ 2.78/ 2.54) | 3.72 | 4.86 | 9.22 | 3.10 |
| | LSM | 9.77 (11.66/ 10.42/ 6.00) | 11.59 | 12.64 | 23.15 | --- |

Correlation coefficient $k_{RT}$ for CFs of group I varies from 0.4 to 0.95. Table 7 details normalized error STD for three intervals of $k_{RT}$. The first of them, $0.4 < k_{RT} < 0.6$, is close to TP#2 considered in Section 3. It is interesting to compare performance of the considered set of estimators on similar simulated and real-life data. For real-life data, performance of the NGF, MI, and NCC estimators are in coherence with the results obtained for pure fBm data: the RST parameters estimation error is 3…7 times greater than $\mathbf{C}_{\boldsymbol{\theta}}$ (see Fig. 2). On simulated data, the $ML_{fBm}$ performed very closely to $\mathbf{C}_{\boldsymbol{\theta}}$ at TP#2; for real-life data its performance decreased by a factor of 1.3…1.7 due to deviation of real-life textures from the fBm model. For high correlation, $0.8 < k_{RT} < 0.95$, the $ML_{fBm}$ still shows the best performance, but its gain is less pronounced (for rotation angle and scaling factor, the MI and NGF show performance similar to the $ML_{fBm}$).

We have also tested a more challenging pair of images to register with different spatial resolutions. We kept the same Hyperion band with 30m resolution as the reference image. Landsat 8 band B8 with spatial resolution 15m was used as template image. To simplify the experiment, we have used the same settings as previously but corrected them taking into account the other spatial resolution of the B8 Landsat 8 band. Comparison was restricted to CFs of group I and the two estimators $ML_{fBm}$ and NCC. This last experiment did not show any significant difference in results for the NCC as compared to those presented and discussed above for a pair of images with the same spatial resolution. Performance of the $ML_{fBm}$ in terms of the normalized error STD degraded by a factor of 1.3…1.4. We explain this by the fact that different widths of point spread function for the reference and template images are not taken into account in our model. Nevertheless, in these challenging settings, the $ML_{fBm}$ still outperforms the NCC. This confirms that both the $ML_{fBm}$ estimator and $\mathbf{C}_{\boldsymbol{\theta}}$ bound are robust enough to significant changes in spatial resolution between reference and template images.

Finally, let us consider the Hyperion and Landsat 8 registration accuracy achievable by the SURF method [41] using OpenSURF library [44]. To facilitate the processing, Landsat 8 image was first transformed to Hyperion coordinate system (using $\mathbf{A}_{\mathrm{HtoL.fine}}$ and $\mathbf{d}_{\mathrm{HtoL.fine}}$) and then cropped. The OpenSURF algorithm has found 2321 control points. The MADs of absolute errors calculated over the 300 best control points take the following values: 3.44 pixels in across-track direction ($\Delta s$) and 0.87 pixels in along-track direction ($\Delta t$). These values exceed significantly the MAD of absolute errors for all estimators used in comparison that vary from 0.2 to 0.65 pixels irrespectively from direction.

Based on the experiments and analysis carried out in this paper, we can conclude that while being applied to register real-life multitemporal data, the $ML_{fBm}$ estimator provides smaller absolute and normalized errors as well as a reduced number of outliers as compared to the state-of-the-art alternative algorithms considered here.

## 5. Discussion

In developing the $ML_{fBm}$ estimator, we have pursued the main goal of improving the image registration accuracy paying attention to the multitemporal and/or multimodal case. The gain obtained with the $ML_{fBm}$ estimator is a 1.75…2 times decrease of the RST parameters estimation error STD and a decrease of false match probability from 50 to 10%. To reach such performance characteristics, we have mainly restricted ourselves to isotropic textures with normal increments that can be described well by the fractional Brownian motion model. We have also dropped willingly computational efficiency requirement. Let us analyze the constraints induced by these assumptions. For real-life data, we have found that about 15% of CFs pass both isotropy and normality tests. Computational burden of the $ML_{fBm}$ is about 35 times higher (for template CFs of 15 by 15 pixels) than that of the NCC as it deals with calculation of full correlation matrices of the registered image fragments.

Fine registration of Lansat8 to Hyperion images (processing nine initial guesses for each of 416 CFs of group I) by the $ML_{fBm}$ implemented in Matlab on Intel Core i7 980X processor takes about 3 hours (rough estimate). Efficient implementation on, for example, C++ programming language could reduce this value by about two times. Registration of individual pairs of CFs is an independent task that can be carried out in parallel. Further decrease can be reached by implementing the basic $ML_{fBm}$ operations with correlation matrices (formation, multiplication, inversion) on GPU. Thus, with an optimization of the implementation, practical registration tasks can be solved with the $ML_{fBm}$ in acceptable time.

Requirement of texture isotropy is quite natural in image registration. Indeed, for anisotropic textures, it becomes impossible to estimate both translation components, but only a linear combination of them. As a result, all analyzed estimators show increased absolute error of the RST parameters estimates for group II (Normal but Anisotropic textures) as compared to group I (Normal and Isotropic textures) and the same tendency for group IV (not Normal and Anisotropic textures) as compared to group III (not Normal but Isotropic textures).

The normality requirement can be justified by the following arguments. First, universal similarity measures like NCC, MI and NGF do not possess robustness for non-normal textures. The drop in accuracy for groups III and IV as compared to groups I and II (see Table 5-7) is as significant for the NCC, MI and NGF estimators as for the $ML_{fBm}$ estimator. Therefore, image registration based on textures with complex structure not following normal distribution is a challenging case for state-of-the-art methods. It requires more efforts to be understood.

Second, assuming normal texture distribution allows formulating image registration problem in terms of second-order statistics. For such a statement, the lower bound of the RST parameters estimation error (CRLB) was derived in closed form (allowing the $ML_{fBm}$ to be viewed as an interval estimator). To the best of our knowledge, this is the only solution that captures the RST parameters estimation error as a function of the texture roughness, reference and template CFs signal-to-noise ratio, correlation between reference and template CFs and RST transformation parameters. Experiments show that this bound is very accurate for both simulated and real data. This bound - an extra outcome of the new estimator we have proposed - can be useful for preliminary detection of CFs suitable for registration, for weighted estimation of global geometrical transformation parameters, or for outlier detection.

Therefore, the $ML_{fBm}$ estimator provides significant advantages over the state-of-the-art RST parameter estimators by introducing natural constraints on image texture but these advantages are gained at the expense of increased computational complexity.

## 6. CONCLUSIONS

This paper presents a new area-based image registration method, $ML_{fBm}$, under rotation-scaling-translation transformation hypothesis.

Experiments on synthetic pure fBm and real hyperspectral data have demonstrated that the $ML_{fBm}$ estimator provides significant decrease in estimation error of the RST transformation parameters as compared to the set of state-of-the-art estimators retained in our comparison. The $ML_{fBm}$ is the most effective estimator in the case of weak correlation between registered CFs (correlation between reference and template images as weak as 0.4…0.6 is acceptable). It has proved to outperform the algorithm based on Mutual Information similarity measure, specially designed to cope with this case.

One interesting feature of the $ML_{fBm}$ is that it provides a CRLB $\mathbf{C}_{\boldsymbol{\theta}}$ on the RST parameters estimation accuracy. For simulated fBm data, the $ML_{fBm}$ error STD is only 1.1 times larger than $\mathbf{C}_{\boldsymbol{\theta}}$. Dealing with complex multitemporal registration of Hyperion and Landsat 8 data, the $ML_{fBm}$ error STD is about 1.5…2 times larger than $\mathbf{C}_{\boldsymbol{\theta}}$. This means that the $ML_{fBm}$ estimator is actually able to provide not only an estimate of the RST transformation vector but also quite an accurate confidence interval for it.

There are two main restrictive features of the $ML_{fBm}$ estimator. First, it relies on the fBm model that might be inadequate when applied to real-life data. Specifically, anisotropic textures, neighborhood of edges, non-random textures, non-Gaussian textures affect its performance. One interesting direction of further studies is to use more complex texture models within the proposed estimation scheme (for example, anisotropic texture models).

The second restrictive feature is that the $ML_{fBm}$ estimator is computationally intensive and, at present, it can be recommended only for "off-line" applications where accuracy is of primary concern.

But the $ML_{fBm}$ estimator (along with the $\mathbf{C}_{\boldsymbol{\theta}}$ bound) has great potential for further development. It can be straightforwardly applied to images formed on irregular grids (for example, due to scanning geometry or relief influence). A more complex affine transformation can be considered as well. Future work will focus on these cases in the framework of multimodal registration.

## APPENDIX A

This appendix defines partial derivatives of the correlation matrix

$$\mathbf{R}_{\Sigma} = \begin{pmatrix} \mathbf{R}_{\mathrm{RI}} + \mathbf{R}_{n.\mathrm{RI}} & k_{\mathrm{RT}} \cdot \mathbf{R}_{\mathrm{RT}}(\boldsymbol{\theta}_{\mathrm{RST}}) \\ k_{\mathrm{RT}} \cdot \mathbf{R}_{\mathrm{RT}}^{T}(\boldsymbol{\theta}_{\mathrm{RST}}) & \mathbf{R}_{\mathrm{TI}} + \mathbf{R}_{\mathrm{n.TI}} \end{pmatrix} = \begin{pmatrix} \mathbf{R}_{\mathrm{RI}} + \mathbf{R}_{n.\mathrm{RI}} & k_{\mathrm{RT}}\sigma_{x.\mathrm{RI}}\sigma_{x.\mathrm{TI}}\mathbf{R}_{\mathrm{HRT}}(\boldsymbol{\theta}_{\mathrm{RST}}) \\ k_{\mathrm{RT}}\sigma_{x.\mathrm{RI}}\sigma_{x.\mathrm{TI}}\mathbf{R}_{\mathrm{HRT}}^{T}(\boldsymbol{\theta}_{\mathrm{RST}}) & \mathbf{R}_{\mathrm{TI}} + \mathbf{R}_{\mathrm{n.TI}} \end{pmatrix}$$

with respect to elements of the parameter vector $\boldsymbol{\theta} = (\sigma_{x.\mathrm{RI}}, \sigma_{x.\mathrm{TI}}, H, k_{\mathrm{RT}}, \Delta t, \Delta s, \alpha, \Delta r)$. The first four derivatives of the matrix $\mathbf{R}_{\Sigma}$ are given in [23]. The derivatives of $\mathbf{R}_{\Sigma}$ w.r.t $\boldsymbol{\theta}_{\mathrm{RST}}$ elements $\Delta t, \Delta s, \alpha, \Delta r$ take the form

$$\frac{\partial \mathbf{R}_{\Sigma}}{\partial \theta_{\mathrm{RST}}(p)} = k_{\mathrm{RT}} \sigma_{x.\mathrm{RI}} \sigma_{x.\mathrm{TI}} \begin{pmatrix} \mathbf{Z}_{\mathrm{RI}} & \dfrac{\partial \mathbf{R}_{\mathrm{HRT}}}{\partial \theta_{\mathrm{RST}}(p)} \\ \left( \dfrac{\partial \mathbf{R}_{\mathrm{HRT}}}{\partial \theta_{\mathrm{RST}}(p)} \right)^T & \mathbf{Z}_{\mathrm{TI}} \end{pmatrix}, \; p = 1...4,$$

where $\mathbf{Z}_{\mathrm{RI}}$ and $\mathbf{Z}_{\mathrm{TI}}$ are $N_{\mathrm{RI}} \times N_{\mathrm{RI}}$ and $N_{\mathrm{TI}} \times N_{\mathrm{TI}}$ zero matrices, respectively. We first give in details the derivation of $R_{\mathrm{HRT}}(k,l)$. We define element $R_{\mathrm{HRT}}(k,l)$ as $R_{\mathrm{HRT}}(k,l) = \left\langle \left( \left[ x(t_k, s_k) - x(0,0) \right] \left[ x(t_l^{'}, s_l^{'}) - x(t_0^{'}, s_0^{'}) \right] \right) \right\rangle$ for $\sigma_{x.\mathrm{RI}} = \sigma_{x.\mathrm{TI}} = 1$ and $k_{\mathrm{RT}} = 1$. Here $(t_l^{'}, s_l^{'})$ and $(t_0^{'}, s_0^{'})$ are coordinates of $(u_l, v_l)$ and $(0,0)$, respectively, in the reference coordinate system obtained according to (1). According to the definition of fBm-process (we refer a reader to [23] for more details on the correlation properties of the fBm model) $R_{\mathrm{HRT}}(k,l)$ can be represented as

$$R_{\mathrm{HRT}}(k,l) = \left\langle \left( \left[ x(t_k, s_k) - x(0,0) \right] \left[ \left( x(t_l^{'}, s_l^{'}) - x(0,0) \right) - \left( x(t_0^{'}, s_0^{'}) - x(0,0) \right) \right] \right) \right\rangle =$$

$$\left\langle \left( \left[ x(t_k, s_k) - x(0,0) \right] \left[ x(t_l^{'}, s_l^{'}) - x(0,0) \right] \right) \right\rangle + \left\langle \left( \left[ x(t_k, s_k) - x(0,0) \right] \left[ x(t_0^{'}, s_0^{'}) - x(0,0) \right] \right) \right\rangle.$$

Using the properties of fBm-process, we finally get:

$$\begin{aligned} R_{\mathrm{HRT}}(k,l) &= \frac{\Delta r^H}{2} \left[ \left( t_k^2 + s_k^2 \right)^H + \left( t_l^{'2} + s_l^{'2} \right)^H - \left( (t_k - t_l^{'})^2 + (s_k - s_l^{'})^2 \right)^H \right] - \\ &- \frac{\Delta r^H}{2} \left[ \left( t_k^2 + s_k^2 \right)^H + \left( t_0^{'2} + s_0^{'2} \right)^H - \left( (t_k - t_0^{'})^2 + (s_k - s_0^{'})^2 \right)^H \right] = \\ &= \frac{\Delta r^H}{2} \left[ \left( (t_k - t_0^{'})^2 + (s_k - s_0^{'})^2 \right)^H + \left( t_l^{'2} + s_l^{'2} \right)^H - \left( t_0^{'2} + s_0^{'2} \right)^H - \left( (t_k - t_l^{'})^2 + (s_k - s_l^{'})^2 \right)^H \right]. \end{aligned}$$

For $p$=1, 2 and 3 (parameters $\Delta t, \Delta s, \alpha$), $\dfrac{\partial R_{\mathrm{HRT}}(k,l)}{\partial \theta_{\mathrm{RST}}(p)}$ is given by

$$\frac{\partial R_{\mathrm{HRT}}(k,l)}{\partial \theta_{\mathrm{RST}} p)} = H \Delta r^H \begin{bmatrix} -\left( (t_k - t_0^{'})^2 + (s_k - s_0^{'})^2 \right)^{H-1} \left( \dfrac{\partial t_0^{'}}{\partial \theta_{\mathrm{RST}}(p)} (t_k - t_0^{'}) + \dfrac{\partial s_0^{'}}{\partial \theta_{\mathrm{RST}}(p)} (s_k - s_0^{'}) \right) + \\ + \left( t_l^{'2} + s_l^{'2} \right)^{H-1} \left( \dfrac{\partial t_l^{'}}{\partial \theta_{\mathrm{RST}}(p)} t_l^{'} + \dfrac{\partial s_l^{'}}{\partial \theta_{\mathrm{RST}}(p)} s_l^{'} \right) - \\ - \left( t_0^{'2} + s_0^{'2} \right)^{H-1} \left( \dfrac{\partial t_0^{'}}{\partial \theta_{\mathrm{RST}}(p)} t_0^{'} + \dfrac{\partial s_0^{'}}{\partial \theta_{\mathrm{RST}}(p)} s_0^{'} \right) + \\ + \left( (t_k - t_l^{'})^2 + (s_k - s_l^{'})^2 \right)^{H-1} \left( \dfrac{\partial t_l^{'}}{\partial \theta_{\mathrm{RST}}(p)} (t_k - t_l^{'}) + \dfrac{\partial s_l^{'}}{\partial \theta_{\mathrm{RST}}(p)} (s_k - s_l^{'}) \right). \end{bmatrix} \qquad (13)$$

For $p$=4 ($\Delta r$ parameter), the term $H\Delta r^{H-2}R_{\mathrm{HRT}}(k,l)$ is added to (13).

Using (1), the derivatives of $t'$, $s'$, $t_0'$ and $s_0'$ with respect to elements $\Delta t, \Delta s, \alpha, \Delta r$ are obtained as:

$$\frac{\partial t'}{\partial \Delta t} = \frac{\partial t_0'}{\partial \Delta t} = -\Delta r^{-1}\cos\alpha\,;\ \frac{\partial t'}{\partial \Delta s} = \frac{\partial t_0'}{\partial \Delta s} = \Delta r^{-1}\sin\alpha\,;\ \frac{\partial t'}{\partial \Delta r} = \frac{\partial t_0'}{\partial \Delta r} = -\Delta r^{-1}t'\,;$$

$$\frac{\partial t'}{\partial \alpha} = -\Delta r^{-1}\left(\sin\alpha\left(u-\Delta t\right)+\cos\alpha\left(v-\Delta s\right)\right);\ \frac{\partial t_0'}{\partial \alpha} = \Delta r^{-1}\left(\Delta t\sin\alpha+\Delta s\cos\alpha\right);$$

$$\frac{\partial s'}{\partial \Delta t} = \frac{\partial s_0'}{\partial \Delta t} = -\Delta r^{-1}\sin\alpha\,;\ \frac{\partial s'}{\partial \Delta s} = \frac{\partial s_0'}{\partial \Delta s} = -\Delta r^{-1}\cos\alpha\,;\ \frac{\partial s'}{\partial \Delta r} = \frac{\partial s_0'}{\partial \Delta r} = -\Delta r^{-1}s'\,;$$

$$\frac{\partial s'}{\partial \alpha} = \Delta r^{-1}\left(\cos\alpha\left(u-\Delta t\right)-\sin\alpha\left(v-\Delta s\right)\right);\ \frac{\partial s_0'}{\partial \alpha} = -\Delta r^{-1}\left(\Delta t\cos\alpha-\Delta s\sin\alpha\right).$$